\documentclass[10pt,twocolumn,letterpaper]{article}

\usepackage{cvpr}

\usepackage{times}
\usepackage{helvet}
\usepackage{courier}
\usepackage[hyphens]{url}
\usepackage{graphicx}
\usepackage{amsmath, amssymb}
\usepackage{booktabs}
\usepackage{caption}
\usepackage{subcaption}
\usepackage{float}
\usepackage{natbib}

\definecolor{cvprblue}{rgb}{0.21,0.49,0.74}
\usepackage[pagebackref,breaklinks,colorlinks,citecolor=cvprblue,linkcolor=cvprblue,urlcolor=magenta]{hyperref}
\usepackage{multirow}
\usepackage{makecell}
\usepackage{xcolor}
\usepackage{colortbl}
\usepackage{bm}
\usepackage[accsupp]{axessibility}

\def\paperID{11293}
\def\confName{CVPR}
\def\confYear{2026}

\title{ParTY: Part-Guidance for Expressive Text-to-Motion Synthesis}

\author{
KunHo Heo \quad
SuYeon Kim \quad
Yonghyun Gwon \quad
Youngbin Kim \quad
MyeongAh Cho$^{\dagger}$ \vspace{0.5em} \\
Kyung Hee University \\
\small\texttt{\{hkh7710, spoiuy3, mathewgwon, youngbean, maycho\}@khu.ac.kr} \vspace{0.2em} \\
\small\href{https://visualsciencelab-khu.github.io/ParTY_project/}{\texttt{https://heokunho.github.io/ParTY/}}
}

\begin{document}
\maketitle
\footnotetext{\dag\ Corresponding author}
\begin{abstract}
Text-to-motion synthesis aims to generate natural and expressive human motions from textual descriptions. While existing approaches primarily focus on generating holistic motions from text descriptions, they struggle to accurately reflect actions involving specific body parts. Recent part-wise motion generation methods attempt to resolve this but face two critical limitations: (i) they lack explicit mechanisms for aligning textual semantics with individual body parts, and (ii) they often generate incoherent full-body motions due to integrating independently generated part motions. To overcome these issues and resolve the fundamental trade-off in existing methods, we propose \textbf{ParTY}, a novel framework that enhances part expressiveness while generating coherent full-body motions. ParTY comprises: \textbf{(1) Part-Guided Network}, which first generates part motions to obtain part guidance, then uses it to generate holistic motions; \textbf{(2) Part-aware Text Grounding}, which diversely transforms text embeddings and appropriately aligns them with each body part; and \textbf{(3) Holistic-Part Fusion}, which adaptively fuses holistic motions and part motions. Extensive experiments, including \textbf{part-level} and \textbf{coherence-level} evaluations, demonstrate that ParTY achieves substantial improvements over previous methods.
\end{abstract}    
\vspace{-6pt}
\section{Introduction}
\label{sec:introduction}

\begin{figure}
    \centering
    \includegraphics[width=1\linewidth]{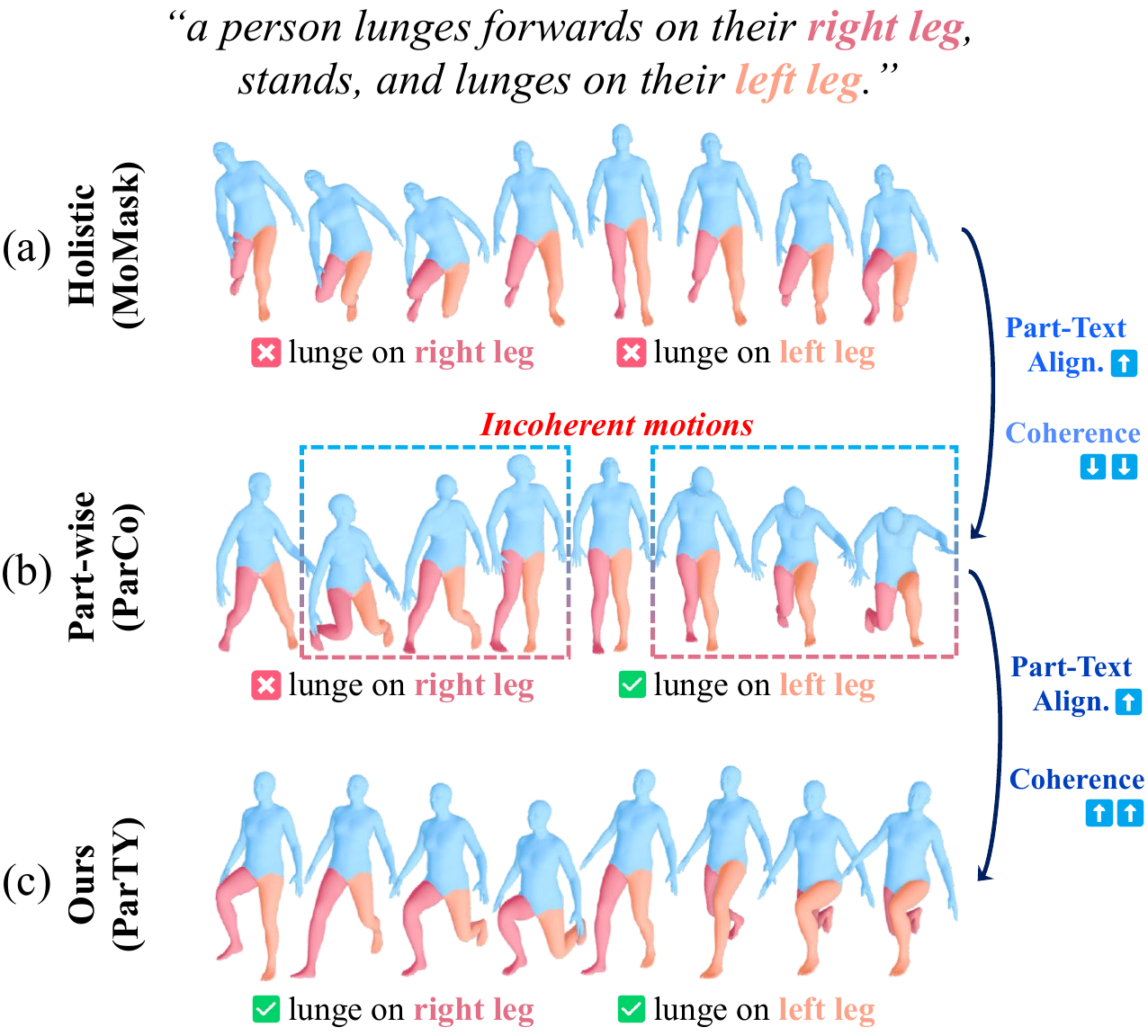} \vspace{-15pt}
    \caption{(a) Holistic methods maintain coherence well but limited part-text alignment. In contrast, (b) Part-wise methods show enhanced part-text alignment (e.g., correctly performing the left leg lunge) but compromised coherence as a trade-off (e.g., neck distortion and misaligned arm and leg movements). (c) Our ParTY resolves this trade-off by achieving superior performance in both part-text alignment and coherence.}
    \vspace{-3pt}
    \label{fig:figure1}
\end{figure}

Text-to-motion synthesis aims to generate human motions from textual descriptions, with applications in animation~\cite{kappel2021high}, virtual reality~\cite{guo2022generating}, video games~\cite{majoe2009enhanced, yeasin2004multiobject}, and robotics~\cite{antakli2018intelligent, koppula2013learning, koppula2015anticipating}. Recent architectures~\cite{zhang2023generating, zhong2023attt2m, guo2024momask, pinyoanuntapong2024mmm, pinyoanuntapong2024bamm, hosseyni2025bad} have improved semantic alignment with the input text, and visual fidelity. However, most adopt a holistic generation approach: synthesizing full-body motion directly from text. While this strategy generates globally coherent motion, it fundamentally lacks the capacity to model part-specific semantics by treating the body as a single entity. As a result, fine-grained part actions in the text are often misrepresented or overlooked, as shown in Fig.~\ref{fig:figure1} (a).

To address this limitation, part-wise methods~\cite{zou2024parco, sun2024lgtm} have emerged, which split the body into anatomical parts and independently generate motions separately for each part. This decomposition provides explicit part-level control and opens up significant potential for improved part-specific expressiveness compared to holistic approaches. However, existing part-wise methods have yet to fully realize this potential and face two critical challenges: \textbf{(i) Insufficient text-to-part semantic alignment}: Existing methods miss fine-grained, part-relevant cues in text, so motions fail to reflect intended part-level behaviors. \textbf{(ii) Lack of inter-part coherence}: Since each part motion is generated independently and then simply combined without consideration of global consistency, the resulting full-body motion often lacks overall coherence, as illustrated by the part-wise method in Fig.~\ref{fig:figure1} (b).

To tackle these challenges and bridge holistic and part-wise methods, we propose \textbf{ParTY}, a novel framework that resolves the inherent trade-off between part-specific expressiveness and full-body coherence. For fine-grained \textit{\textbf{text-to-part alignment}}, we introduce \textbf{Part-aware Text Grounding} module: it transforms a single sentence embedding into multiple diverse embeddings and dynamically selects appropriate embeddings for each body part. During the selection process, it leverages auxiliary text information about each part generated by an LLM, which is used only during training. This enables fine-grained text-part alignment over previous methods.

To address the lack of \textit{\textbf{coherence}} among independently generated part motions, we propose a \textbf{Part-Guided Network}, a dual-generation framework that first generates part motions and then uses them as guidance to generate holistic motions, rather than generating each part independently and combining them. Specifically, part motions are generated for several time steps to create part guidance, which conditions the holistic motion generation by providing future part-level information. During holistic motion generation, a \textbf{Holistic-Part Fusion} is also employed, which directly fuses holistic and part motions, allowing part motion information to be incorporated throughout the process. These approaches enable the generation of coherent movements across the entire body.

While our method improves part-specific expressiveness and full-body coherence, evaluating these improvements remains challenging. Conventional metrics~\cite{guo2022generating} operate exclusively at the holistic-level, making them incapable of accurately assessing part-level semantic alignment. Moreover, no metric exists to directly evaluate motion coherence across the full-body. To address these issues, we propose new evaluation protocols: \textbf{part-level metrics} that expand~\cite{guo2022generating}'s approach for evaluating part-specific expressiveness and \textbf{temporal and spatial coherence metrics}. Extensive experiments including these evaluations and Fig.~\ref{fig:figure1} (c) demonstrate that our \textbf{ParTY} effectively combines the advantages of both holistic and part-wise methods, leveraging expressive part motions while maintaining coherence.

\noindent Our contributions are summarized as follows:
\begin{itemize}
    \item We introduce a \textbf{Part-Guided Network} that addresses the \textit{coherence} problem inherent to part-wise methods by first generating part motions and then using them as guidance for holistic motion generation, with a \textbf{Holistic-Part Fusion} module that adaptively fuses both motion representations to maintain coordination.
    \item We propose \textbf{Part-aware Text Grounding}, which enhances fine-grained \textit{text-to-part alignment} by diversely transforming text embeddings and appropriately aligning them with each body part, leveraging LLM-generated part descriptions as auxiliary information.
    \item Through extensive experiments, we demonstrate that our method achieves state-of-the-art performance on conventional metrics. Furthermore, using our newly proposed \textbf{part-level and coherence-level evaluation metrics}, we validate and analyze the effectiveness of our approach in improving part expressiveness and motion coherence.
\end{itemize}
\section{Related Works}
\label{sec:related_work}

\textbf{Text-to-Motion Generation.} Text-to-motion synthesis aims to translate textual descriptions into realistic human movements, offering intuitive control through natural language~\cite{bhattacharya2021text2gestures, plappert2018learning}. Unlike traditional methods relying on action classes~\cite{lucas2022posegpt, petrovich2021action, guo2020action2motion} or audio signals~\cite{lee2019dancing, li2021ai, tseng2023edge}, text-driven approaches provide expressive control capabilities. Early approaches established cross-modal alignment through joint embedding spaces~\cite{ahuja2019language2pose, ahn2018text2action}, followed by probabilistic frameworks~\cite{petrovich2022temos, athanasiou2022teach} that captured the one-to-many relationship between text and motion. Discrete representation learning emerged with~\cite{guo2022tm2t} enhancing cross-modal understanding, while~\cite{zhang2023generating, jiang2023motiongpt} combined VQ-VAE~\cite{van2017neural} with transformers for autoregressive generation. Recent diffusion-based methods~\cite{chen2023executing, gao2024guess, tevet2023human, wang2023fg, zhang2022motiondiffuse} dramatically improve motion quality through iterative denoising processes, though often with computational overhead. Temporal context modeling has been explored through masked modeling strategies~\cite{guo2024momask} and hybrid approaches~\cite{pinyoanuntapong2024bamm, pinyoanuntapong2024mmm}. Despite these advances, most approaches treat human motion as a monolithic entity rather than a coordinated system of interrelated parts, struggling to capture nuanced relationships between textual descriptions and specific body movements that require precise inter-part coordination.

\noindent \textbf{Part-wise Text-to-Motion Generation.} Part-wise approaches treat the human body as a system of coordinated parts rather than a monolithic entity, aiming to enhance expressiveness and control over individual body components. Early work included SCA~\cite{ghosh2021synthesis}, which divided the body into upper and lower segments with independent networks, demonstrating potential for specialized control but struggling with coordination. Additionally, AttT2M~\cite{zhong2023attt2m} introduced body-part attention-based encoders, though its single decoder limited nuanced control. More recently, ParCo~\cite{zou2024parco} used separate VQ-VAEs for each body part with token-sharing for coordination, but did not leverage part-specific text descriptions. LGTM~\cite{sun2024lgtm} decomposed text descriptions into part-specific prompts using an LLM, but this extraction of only part-related text loses the overall context of the sentence. Due to these limitations, although these approaches achieve some improvement in part expressiveness, they still exhibit insufficient part-level detail. Moreover, the simple integration of independently generated parts inevitably results in incoherent motions. Our method addresses these limitations, enhancing both aspects simultaneously.

\begin{figure}
    \centering
    \includegraphics[width=1\linewidth]{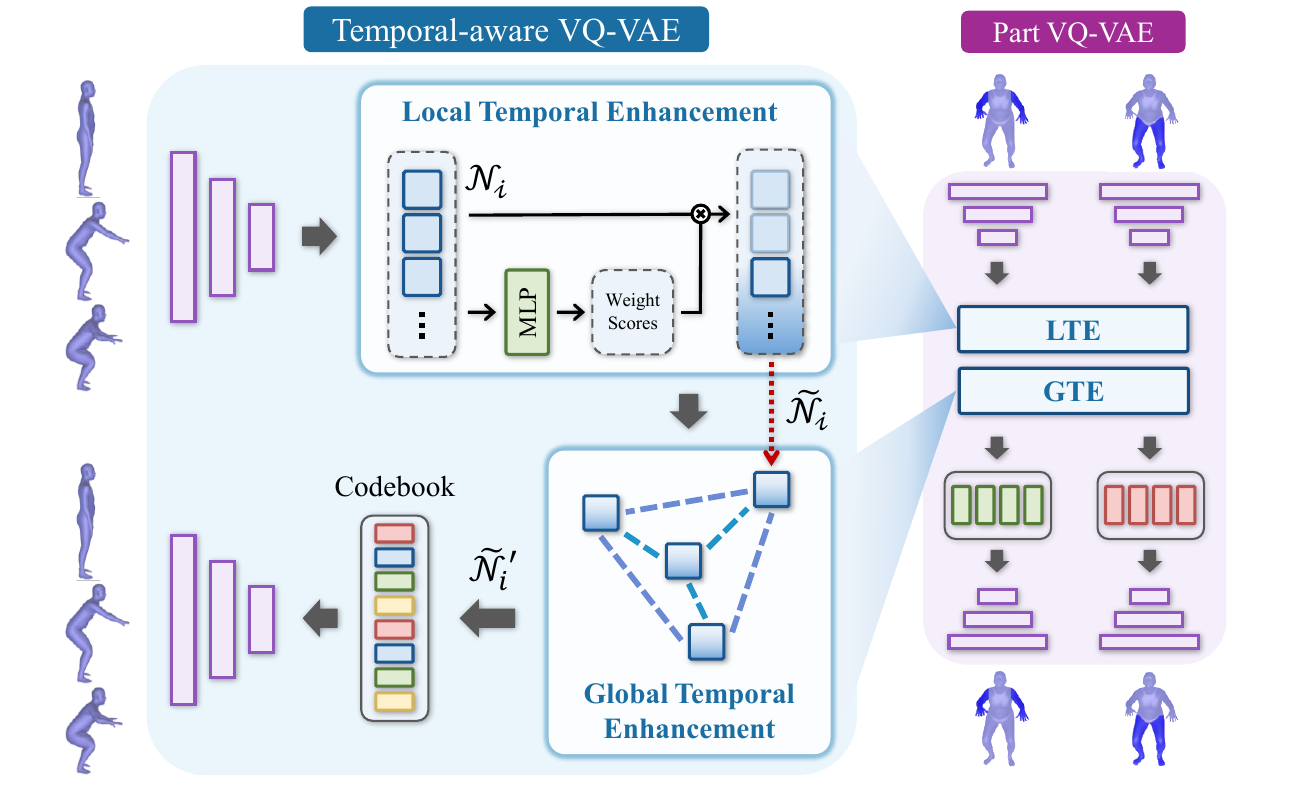}
    \caption{Architecture of the Temporal-aware VQ-VAE. Part VQ-VAE follows an identical architecture, where the sole distinction lies in processing part-level rather than full-body motion data.}
    \label{fig:figure2}
\end{figure}
\begin{figure*}[t!]
    \centering
    \includegraphics[width=\linewidth]{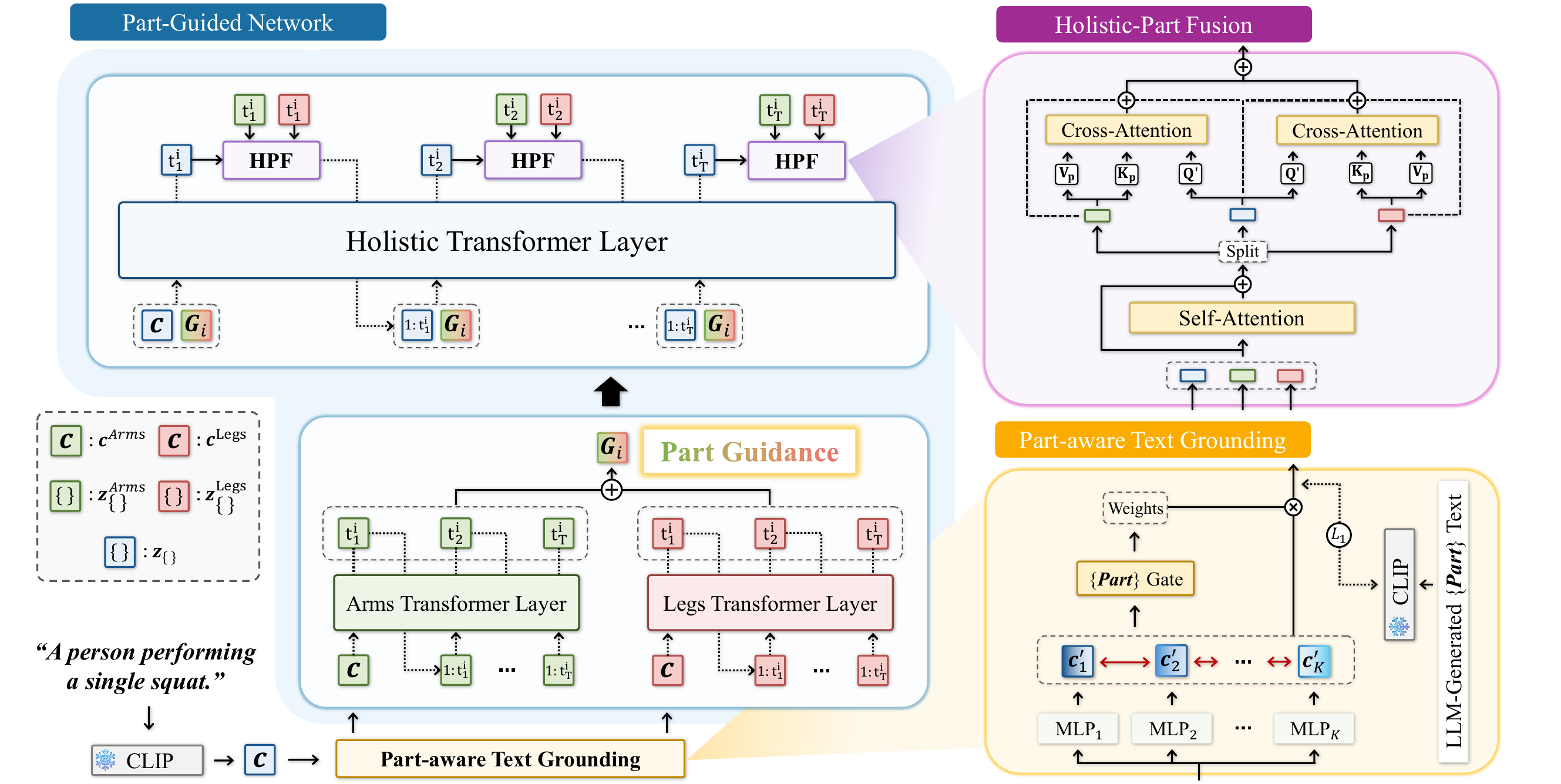}
    \caption{Overview of ParTY. Text embeddings are processed through Part-aware Text Grounding, then part transformers generate Part Guidance for the holistic transformer to generate motion tokens, with Holistic-Part Fusion applied during generation. The notation \{\textbf{\textit{Part}}\} indicates that the process is performed for both arms and legs.}
    \label{fig:figure3}
\end{figure*}

\section{Method}
Our method consists of two stages. First, we quantize motion sequences into codebooks for both the full-body and parts (arms and legs), as shown in Fig.~\ref{fig:figure2}. Second, we train holistic and part transformers using these codebooks to predict codebook sequences that match the text description, as shown in Fig.~\ref{fig:figure3}. During inference, the predicted codebook sequences are decoded by the pre-trained VQ-VAE decoder from the first stage to reconstruct the motions. Further network details can be found in the supplementary material.

\subsection{Temporal-aware VQ-VAE}
Recent studies have increasingly adopted VQ-VAE~\cite{van2017neural} to quantize sequential motions into discrete codebooks~\cite{zhang2023generating, guo2024momask, pinyoanuntapong2024bamm, pinyoanuntapong2024mmm, zou2024parco, hosseyni2025bad}. However, compressing motion sequences—where temporal flow is crucial—into discrete codebooks through fixed-size windows inherently causes temporal information loss. While reducing the window size can mitigate this loss by increasing codebook entries, this introduces a trade-off with model size without truly resolving the problem. This constitutes a fundamental limitation of VQ-VAE-based approaches. To address this limitation without increasing model complexity, we propose \textbf{Temporal-aware VQ-VAE} that enhances both local and global temporal information to enable codebook quantization while preserving temporal details.

Our goal is to quantize motion sequences into codebooks while preserving temporal information. To achieve this, we enhance local and global temporal information before quantization, as shown in Fig.~\ref{fig:figure2}. As motion sequences are encoded frame-by-frame through the encoder, \textbf{Local Temporal Enhancement (LTE)} bundles frame-level features with a window size of $w$, resulting in feature groups $\{\mathcal{N}_i\}_{i=1}^{t/w}$ where $t$ denotes total frames. For each group, we compute feature weights via an $\text{MLP}_i$ (a 3-layer MLP with ReLU activation) and perform weighted summation to produce an enhanced group-level feature $\{\tilde{\mathcal{N}}_i\}_{i=1}^{t/w}$, formulated as:

\begin{equation}
\tilde{\mathcal{N}}_i = \sum_{j=1}^{w} \alpha_{ij} \cdot f_{ij}, \quad f_{ij} \in \mathcal{N}_i
\end{equation}

\noindent where $f_{ij}$ is the $j$-th feature in group $\mathcal{N}_i$, and $\alpha_{ij}$ is the weight computed by applying softmax to $\text{MLP}_i$ outputs, ensuring $\sum_{j=1}^{w} \alpha_{ij} = 1$.

In \textbf{Global Temporal Enhancement (GTE)}, we employ a Graph Convolutional Network~\cite{kipf2016semi} to preserve global temporal dependencies. We define the nodes of the GCN as the group-level features $\{\tilde{\mathcal{N}}_i\}_{i=1}^{t/w}$ and update each node by capturing relationships among them:

\begin{equation}
\tilde{\mathcal{N}}'_i = \text{GELU}\left(\sum_{k=1}^{t/w} \hat{A}_{ik}(\tilde{\mathcal{N}}_k W)\right)
\end{equation}

\noindent where $\tilde{\mathcal{N}}'_i$ is the updated $i$-th node feature, $\tilde{\mathcal{N}}_k$ is the $k$-th node feature, $\hat{A}_{ik}$ is the normalized adjacency matrix, and $W$ is a learnable weight matrix. Finally, each $\tilde{\mathcal{N}}'_i$ is mapped to a single codebook entry through quantization. By preserving temporal information, our approach encodes longer motion sequences into single codebook entries with reduced information loss, thereby decreasing model size and inference time. Related analysis is provided in Tab.~\ref{tab:table4} and Sec.~\ref{sec:discussions}.

\noindent \textbf{Training.} Our VQ-VAE  is optimized by $\mathcal{L}_{vq} = \mathcal{L}_{rec} + \lambda_{app} \cdot \mathcal{L}_{app}$, where $\mathcal{L}_{rec}$ is the $L_1$ reconstruction loss between decoded and ground truth joint positions, and $\mathcal{L}_{app}$ is the $L_2$ approximation loss between quantized codebook vectors and the encoded vectors, which encourages the encoder to produce features close to learned codebook entries.

\subsection{Part-aware Text Grounding}
After completing the motion quantization stage, in the second stage, we proceed to motion generation using transformers. Before feeding text embedding into each part's transformer, we apply Part-aware Text Grounding (PTG) to generate text embeddings tailored to each body part. As shown in Fig.~\ref{fig:figure3}, we first obtain text embedding $\mathbf{c}$ from the text description through CLIP~\cite{radford2021learning}. This embedding is fed into $K$ distinct MLPs, producing transformed embeddings $\mathbf{c}'_n = \mathrm{MLP}_n(\mathbf{c})_{n \in \{1, \dots, K\}}$. To ensure these embeddings maintain semantic consistency while achieving diversity, we employ contrastive learning where each $\mathbf{c}'_n$ treats $\mathbf{c}$ as a positive anchor and $\{\mathbf{c}'_m \}_{m \neq n}^{K}$ as negatives. The text diversity loss is:
\begin{equation}
\small
\begin{gathered}
\mathcal{L}_{\text{div}} = \frac{1}{K} \sum_{n=1}^{K} \mathcal{L}^{(n)} \\[0.5em]
{\small \mathcal{L}^{(n)} = -\log \frac{\exp(\text{s}(\mathbf{c}'_n, \mathbf{c})/{\tau})}{\exp(\text{s}(\mathbf{c}'_n, \mathbf{c})/{\tau}) + \sum\limits_{m \neq n}^{K} \exp(\text{s}(\mathbf{c}'_n, \mathbf{c}'_m)/{\tau})}}
\end{gathered}
\end{equation}
\noindent where $\text{s}(\cdot, \cdot)$ denotes cosine similarity and ${\tau}$ is the temperature parameter. This encourages each MLP to explore different semantic aspects while preserving core meaning.
After generating diverse embeddings, the \{\textit{\textbf{Part}}\} Gate—a gating network dedicated to arms and legs—dynamically selects appropriate embeddings through adaptive weighting. To improve part-specific selection, we use LLM-Generated \{\textit{\textbf{Part}}\} Text as auxiliary supervision: detailed motion descriptions for each part are generated from the original text (e.g., given \textit{``a person walks forward and picks something up off the ground with their left hand,"} the LLM generates \textit{``Left arm picks something up off the ground"} for the arm and \textit{``The legs step forward one after the other"} for the legs), embedded via CLIP, and used to compute an auxiliary $L_1$ loss $\mathcal{L}_{\text{aux}}$ that aligns PTG outputs with these part-specific embeddings. Notably, since LLM-generated texts are only used during training, our method remains efficient at inference time.

\subsection{Part-Guided Network}
To achieve expressive motion generation, we propose a Part-Guided Network that first generates part motion tokens for a certain number of time steps and leverages them as part guidance for generating holistic motion tokens. Following PTG, the precisely aligned text embeddings are fed into their respective part transformers, which autoregressively generate part motion tokens. The generation proceeds in cycles: in the $i$-th cycle, each part transformer generates $T$ consecutive tokens for time steps $t_i = \{t^i_1, t^i_2, \ldots, t^i_T\}$, and the generated tokens from each part are fused to create the $i$-th Part Guidance $\mathbf{G}_i$:
\begin{gather}
\mathbf{G}_i = \sum_{t \in t_i} \mathbf{z}^{\text{fuse}}_{t} \\
\mathbf{z}^{\text{fuse}}_t = \text{MLP}(\mathbf{z}^{\text{Arms}}_t + \mathbf{z}^{\text{Legs}}_t) \\
\mathbf{z}^p_t = f_p(\mathbf{z}^p_{1:t-1}, \mathbf{c}^p), \quad p \in \{\text{Arms}, \text{Legs}\}
\end{gather}
where $f_p$ denotes part transformer that autoregressively generates motion token $\mathbf{z}^p_t$ conditioned on previous tokens $\mathbf{z}^p_{1:t-1}$ and part-specific text embedding $\mathbf{c}^p$ from PTG.

Next, the $i$-th Part Guidance is fed into the holistic transformer at each step during the generation of holistic motion tokens for time steps in $t_i$. The holistic motion token $\mathbf{z}_t$ generated at time step $t \in t_i$ by the holistic transformer is formulated as:
\begin{gather}
\mathbf{z}_t = f(\mathbf{z}_{1:t-1}, \mathbf{c}, \mathbf{G}_i) \\
\mathbf{z}_{1:t-1} = \texttt{HPF}(\mathbf{z}_{1:t-1}, \mathbf{z}^{\text{Arms}}_{1:t-1},\mathbf{z}^{\text{Legs}}_{1:t-1})
\end{gather}
where $\mathbf{z}_{1:t-1}$ denotes the previous holistic motion tokens refined by the Holistic-Part Fusion (HPF), $\mathbf{c}$ is original text embedding without processing through the PTG module. Overall, this process of the $i$-th generation cycle—generating part motion tokens for $T$ steps and subsequently generating holistic motion tokens for $T$ steps—continues cyclically until the holistic transformer reaches the end token.

\noindent \textbf{Holistic-Part Fusion.} To ensure coherent movements across the entire body, we continuously integrate part motion information into the holistic transformer through the Holistic-Part Fusion (HPF) during holistic motion token generation. HPF first concatenates the holistic tokens $\mathbf{z}_{1:t-1}$ with arm tokens $\mathbf{z}^{\text{Arms}}_{1:t-1}$ and leg tokens $\mathbf{z}^{\text{Legs}}_{1:t-1}$, then performs self-attention using the standard scaled dot-product attention mechanism $\text{Attn}(\mathbf{Q}, \mathbf{K}, \mathbf{V}) = \text{softmax}\left(\frac{\mathbf{Q}\mathbf{K}^\top}{\sqrt{d_k}}\right)\mathbf{V}$, where $\mathbf{Q}$, $\mathbf{K}$, $\mathbf{V}$ are linear projections of the concatenated tokens [$\mathbf{z}_{1:t-1}$; $\mathbf{z}^{\text{Arms}}_{1:t-1}$; $\mathbf{z}^{\text{Legs}}_{1:t-1}$]. The attended output is then split back into separate sequences $\tilde{\mathbf{z}}_{1:t-1}$, $\tilde{\mathbf{z}}^{\text{Arms}}_{1:t-1}$, and $\tilde{\mathbf{z}}^{\text{Legs}}_{1:t-1}$ using split tokens, which are learnable vectors that serve as separators between different tokens. We apply cross-attention operations with $\tilde{\mathbf{z}}_{1:t-1}$ as query and each part token as key/value:
\begin{align}
\small
\mathbf{z}^{\text{p}}_{\text{cross}} &= \text{Attn}(\mathbf{Q}', \mathbf{K}_{\text{p}}, \mathbf{V}_{\text{p}}), \quad p \in \{\text{Arms}, \text{Legs}\} 
\end{align}
\noindent where $\mathbf{Q}'$ and $(\mathbf{K}_{\text{p}}, \mathbf{V}_{\text{p}})$ are linear projections of $\tilde{\mathbf{z}}_{1:t-1}$ and $\tilde{\mathbf{z}}^{\text{p}}_{1:t-1}$, respectively. Finally, the two cross-attention outputs are added to produce the HPF output.

\noindent \textbf{Training.} To train both holistic and part transformers, we design separate loss functions for each. Let $d(\mathbf{z}|t)$ denote the conditional distribution of $\mathbf{z}$ at time step $t$. The holistic motion loss and part motion loss are defined as:
\begin{gather}
\mathcal{L}_{\text{hol}} = \mathbb{E}_{\mathbf{z}_t, t \sim d(\mathbf{z}_t, t)}[-\log d(\mathbf{z}_t|t)] \\
\mathcal{L}_{\text{part}} = \sum_{p \in \{\text{Arms}, \text{Legs}\}} \mathbb{E}_{\mathbf{z}^p_t, t \sim d(\mathbf{z}^p_t, t)}[-\log d(\mathbf{z}^p_t|t)]
\end{gather}
The total loss is $\mathcal{L} = \mathcal{L}_{\text{hol}} + \mathcal{L}_{\text{part}} + \lambda_{\text{div}} \mathcal{L}_{\text{div}} + \lambda_{\text{aux}} \mathcal{L}_{\text{aux}}$, where $\lambda_{\text{div}}$ and $\lambda_{\text{aux}}$ are weighting coefficients for the text diversity loss and part text auxiliary loss, respectively.
\section{Experiments}
\begin{table*}[t]
\caption{Quantitative comparison on HumanML3D and KIT-ML. \textbf{Bold} indicates the best result, while \underline{underlined} refers the second-best. The right arrow $\rightarrow$ indicates that closer values to ground truth are preferred.}
\vspace{-2mm}
\small
\resizebox{\linewidth}{!}{%
\begin{tabular}{@{}clccccccc@{}}
\toprule
\multirow{2}{*}{Datasets} & \multirow{2}{*}{Method} & \multicolumn{3}{c}{R Precision $\uparrow$}                                                                                                                & \multicolumn{1}{c}{\multirow{2}{*}{FID$\downarrow$}} & \multirow{2}{*}{MM-Dist$\downarrow$}              & \multirow{2}{*}{Diversity$\rightarrow$}           & \multirow{2}{*}{MultiModality$\uparrow$}              \\ \cmidrule(lr){3-5}
             ~& ~& \multicolumn{1}{c}{Top 1} & \multicolumn{1}{c}{Top 2} & \multicolumn{1}{c}{Top 3} & \multicolumn{1}{c}{}                     &                          &                            &                            \\ \midrule
\multirow{8}{*}{\makecell[c]{HumanML3D}}  &  Real motion & 
 $0.511^{\pm.003}$ &
 $0.703^{\pm.003}$ &
 $0.797^{\pm.002}$ &
 $0.002^{\pm.000}$ &
 $2.974^{\pm.008}$ &
 $9.503^{\pm.065}$ &
  - \\
 \cline{2-9} \vspace{-0.3cm} \\ 

~& MDM \cite{tevet2022human}&
  $0.320^{\pm.005}$ &
  $0.498^{\pm.004}$ &
  $0.611^{\pm.007}$ &
  ${0.544}^{\pm.044}$ &
  $5.566^{\pm.027}$ &
  ${\underline{9.559}}^{\pm.086}$ &
  $\mathbf{2.799}^{\pm.072}$ \\

~& T2M-GPT \cite{zhang2023generating} &
    ${0.491}^{\pm.003}$ &
    ${0.680}^{\pm.003}$ &
    ${0.775}^{\pm.002}$ &
    ${0.116}^{\pm.004}$ &
    ${3.118}^{\pm.011}$ &
    ${9.761}^{\pm.081}$ &
    $1.856^{\pm.011}$ \\ 

~& ParCo \cite{zou2024parco} &
    ${0.515}^{\pm.003}$ &
    ${0.706}^{\pm.003}$ &
    ${0.801}^{\pm.002}$ &
    ${0.109}^{\pm.005}$ &
    ${2.927}^{\pm.008}$ &
    ${9.576}^{\pm.088}$ &
    $1.382^{\pm.060}$ \\

~& MMM \cite{pinyoanuntapong2024mmm} &
    ${0.504}^{\pm.003}$ &
    ${0.696}^{\pm.003}$ &
    ${0.794}^{\pm.002}$ &
    ${0.080}^{\pm.003}$ &
    ${2.998}^{\pm.007}$ &
    ${9.411}^{\pm.058}$ &
    $1.164^{\pm.041}$ \\ 

 ~& BAMM \cite{pinyoanuntapong2024bamm} &
    ${\underline{0.525}}^{\pm.002}$ &
    ${\underline{0.720}}^{\pm.003}$ &
    ${\underline{0.814}}^{\pm.003}$ &
    ${0.055}^{\pm.002}$ &
    ${\underline{2.919}}^{\pm.008}$ &
    ${9.717}^{\pm.089}$ &
    $1.687^{\pm.051}$ \\

 ~& MoMask \cite{guo2024momask} &
    ${0.521}^{\pm.002}$ &
    ${0.713}^{\pm.002}$ &
    ${0.807}^{\pm.002}$ &
    ${\underline{0.045}}^{\pm.002}$ &
    ${2.958}^{\pm.008}$ &
    - &
    $1.241^{\pm.040}$ \\

\cline{2-9} \vspace{-0.3cm} \\ 
~& \textbf{ParTY (Ours)} &
    $\mathbf{0.550}^{\pm.003}$ &
    $\mathbf{0.744}^{\pm.003}$ &
    $\mathbf{0.836}^{\pm.003}$ &
    $\mathbf{0.035}^{\pm.002}$ &
    $\mathbf{2.779}^{\pm.006}$ &
    $\mathbf{9.534}^{\pm.066}$ &
    $\underline{2.155}^{\pm.046}$   
    \\ \midrule

\multirow{8}{*}{\makecell[c]{KIT-ML}}  &  Real motion & 
 $0.424^{\pm.005}$ &
 $0.649^{\pm.006}$ &
 $0.779^{\pm.006}$ &
 $0.031^{\pm.004}$ &
 $2.788^{\pm.012}$ &
 $11.08^{\pm.097}$ &
  -
  \\ 
\cline{2-9} \vspace{-0.3cm} \\ 

~& MDM \cite{tevet2022human}&
  $0.164^{\pm.004}$ &
  $0.291^{\pm.004}$ &
  $0.396^{\pm.004}$ &
  ${0.497}^{\pm.021}$ &
  $9.190^{\pm.022}$ &
  ${10.85}^{\pm.109}$ &
  $\mathbf{1.907}^{\pm.214}$ \\

~& T2M-GPT \cite{zhang2023generating} &
    ${0.416}^{\pm.006}$ &
    ${0.627}^{\pm.006}$ &
    ${0.745}^{\pm.006}$ &
    ${0.514}^{\pm.029}$ &
    ${3.007}^{\pm.023}$ &
    ${10.92}^{\pm.108}$ &
    $1.570^{\pm.039}$ \\  

~& ParCo \cite{zou2024parco} &
    ${0.430}^{\pm.004}$ &
    ${0.649}^{\pm.007}$ &
    ${0.772}^{\pm.006}$ &
    ${0.453}^{\pm.027}$ &
    ${2.820}^{\pm.028}$ &
    ${10.95}^{\pm.094}$ &
    $1.245^{\pm.022}$ \\

~& MMM \cite{pinyoanuntapong2024mmm} &
    ${0.404}^{\pm.005}$ &
    ${0.621}^{\pm.005}$ &
    ${0.744}^{\pm.004}$ &
    ${0.316}^{\pm.028}$ &
    ${2.977}^{\pm.019}$ &
    ${10.91}^{\pm.101}$ &
    $1.232^{\pm.039}$ \\ 

 ~& BAMM \cite{pinyoanuntapong2024bamm} &
    ${\underline{0.438}}^{\pm.009}$ &
    ${\underline{0.661}}^{\pm.009}$ &
    ${\underline{0.788}}^{\pm.005}$ &
    ${\underline{0.183}}^{\pm.013}$ &
    ${\underline{2.723}}^{\pm.026}$ &
    $\mathbf{11.01}^{\pm.094}$ &
    $\underline{1.609}^{\pm.065}$ \\

 ~& MoMask \cite{guo2024momask} &
    ${0.433}^{\pm.007}$ &
    ${0.656}^{\pm.005}$ &
    ${0.781}^{\pm.005}$ &
    ${0.204}^{\pm.011}$ &
    ${2.779}^{\pm.022}$ &
    - &
    $1.131^{\pm.043}$ \\

\cline{2-9} \vspace{-0.3cm} \\ 
~& \textbf{ParTY (Ours)} &
    $\mathbf{0.449}^{\pm.006}$ &
    $\mathbf{0.680}^{\pm.007}$ &
    $\mathbf{0.804}^{\pm.006}$ &
    $\mathbf{0.155}^{\pm.014}$ &
    $\mathbf{2.694}^{\pm.030}$ &
    ${\underline{11.21}}^{\pm.082}$ &
    ${1.166}^{\pm.049}$ 
    \\ \bottomrule
\end{tabular}%
}
\label{tab:table1}
\vspace{-1mm}
\end{table*}

We evaluate on HumanML3D~\cite{guo2022generating} (14,616 motions, 44,970 texts) and KIT-ML~\cite{plappert2016kit} (3,911 motions, 6,278 texts), following the standard splits and pose representation from~\cite{guo2022generating}. Additional quantitative and qualitative experiment results can be found in the supplementary material.
\subsection{Evaluation Metrics}
We adopt evaluation metrics from~\cite{guo2022generating} using pre-trained encoders. R-Precision and MM-Dist measure text-motion alignment and semantic similarity in the feature space. FID evaluates motion quality through distributional differences between generated and real motions. We also report Diversity (variance across motion pairs) and Multimodality (variance for motions from the same text). Following prior work, we run each evaluation 20 times (5 times for Multimodality) and report averages with 95\% confidence intervals.

\subsubsection{Part-level Evaluation Metrics}
To evaluate part-specific expressiveness, we extend conventional evaluation metrics to the part-level. We independently train arms and legs motion encoders using the T2M~\cite{guo2022generating} encoder architecture, which is commonly used for evaluation in most studies~\cite{zhang2023generating, zhong2023attt2m, pinyoanuntapong2024mmm, pinyoanuntapong2024bamm, zou2024parco}. With these trained part-specific encoders, we compute part-level R-Precision, FID, and MM-Dist by adapting full-body metrics to evaluate part-level performance.
\subsubsection{Coherence-level Evaluation Metrics}
To evaluate motion coherence at the frame-level, we introduce the Temporal Coherence (TC) and Spatial Coherence (SC) score, which evaluate both temporal and spatial consistency across body parts. A motion sequence is represented by $j$ joints with 3D position $\hat{\mathbf{p}}_j(t)$ at time step $t$, partitioned into five body parts: left arm, right arm, left leg, right leg, and backbone.

\noindent \textbf{Temporal Coherence} score quantifies temporal coordination between body parts over time. For each body part $g$, we compute the temporal-wise RMS velocity:
\vspace{-2pt}
\begin{equation}
\mathbf{x}_g(t) = {\sqrt{\frac{1}{n_g} \sum_{j \in g} \|\hat{\mathbf{p}}_j(t) - \hat{\mathbf{p}}_j(t-1)\|^2}}
\vspace{-1mm}
\end{equation}
\noindent where the sum is over all joints $j$ belonging to part $g$, and $n_g$ is the number of joints from part $g$. To compare motion patterns across parts with different movement intensities, we apply z-normalization within sliding windows $w$—local time intervals that allow adaptive normalization to account for varying motion dynamics throughout the sequence. We then measure part-wise correlations within each window through cross-correlation functions $r_{g,h}^{(w)}(\tau)$ across temporal lags $\tau$. This allows detection of phase-shifted synchrony—for instance, in walking, arm motion naturally leads or lags leg motion. To aggregate these correlations, we compute a refined correlation score:
\vspace{-3pt}
\begin{equation}
\tilde{R}_{g,h}^{(w)} = \max(0, \mathbb{E}_{\tau}[r_{g,h}^{(w)}(\tau)]) \cdot \exp\left(-\frac{\langle |\tau| \rangle_w}{\kappa}\right)
\vspace{-1mm}
\end{equation}
where $\mathbb{E}_{\tau}[\cdot]$ denotes softmax-weighted averaging over lags (emphasizing high-correlation lags), $\langle |\tau| \rangle_w$ is the expected absolute lag, and the exponential term penalizes excessive delays to suppress spurious matches from unrelated movements. The temporal coherence score is then $S_{\text{temporal}} = \frac{1}{W|\mathcal{P}|} \sum_{w=1}^W \sum_{(g,h) \in \mathcal{P}} \tilde{R}_{g,h}^{(w)}$, where $W$ is the number of windows and $\mathcal{P}$ is the set of all part pairs, yielding a measure of overall rhythmic coordination.

\noindent \textbf{Spatial Coherence} score evaluates the physical plausibility of spatial relationships within each frame. For each body part $g$, we compute its representative position as the centroid (average 3D position) of all joints belonging to that part: $\mathbf{c}_g(t) = \frac{1}{n_g} \sum_{j \in g} \hat{\mathbf{p}}_j(t)$. We then measure: (1) inter-part distances $d_{g,h}(t) = \|\mathbf{c}_g(t) - \mathbf{c}_h(t)\|$ between the centroids of part pairs $(g,h)$, and (2) part-torso angles $\theta_g(t)$ measuring the angular alignment between each part and the torso. To define what constitutes \textit{physically plausible} human poses, we estimate reference statistics—means and standard deviations—of these quantities over the HumanML3D dataset~\cite{guo2022generating}, forming empirical distributions of natural human geometry. Each per-frame measurement is normalized into z-scores $z_{g,h}^{(d)}$ and $z_g^{(\theta)}$, which are converted to consistency scores using Gaussian kernels:
\begin{equation}
\small
s_{g,h}^{(d)}(t) = \exp\left(-\frac{(z_{g,h}^{(d)}(t))^2}{\beta_d^2}\right), \quad s_g^{(\theta)}(t) = \exp\left(-\frac{(z_g^{(\theta)}(t))^2}{\beta_\theta^2}\right)
\end{equation}
\noindent where larger deviations from typical human geometry lead to exponentially lower consistency. The spatial coherence score is defined as:
\begin{figure*}[t!]
    \centering
    \includegraphics[width=\linewidth]{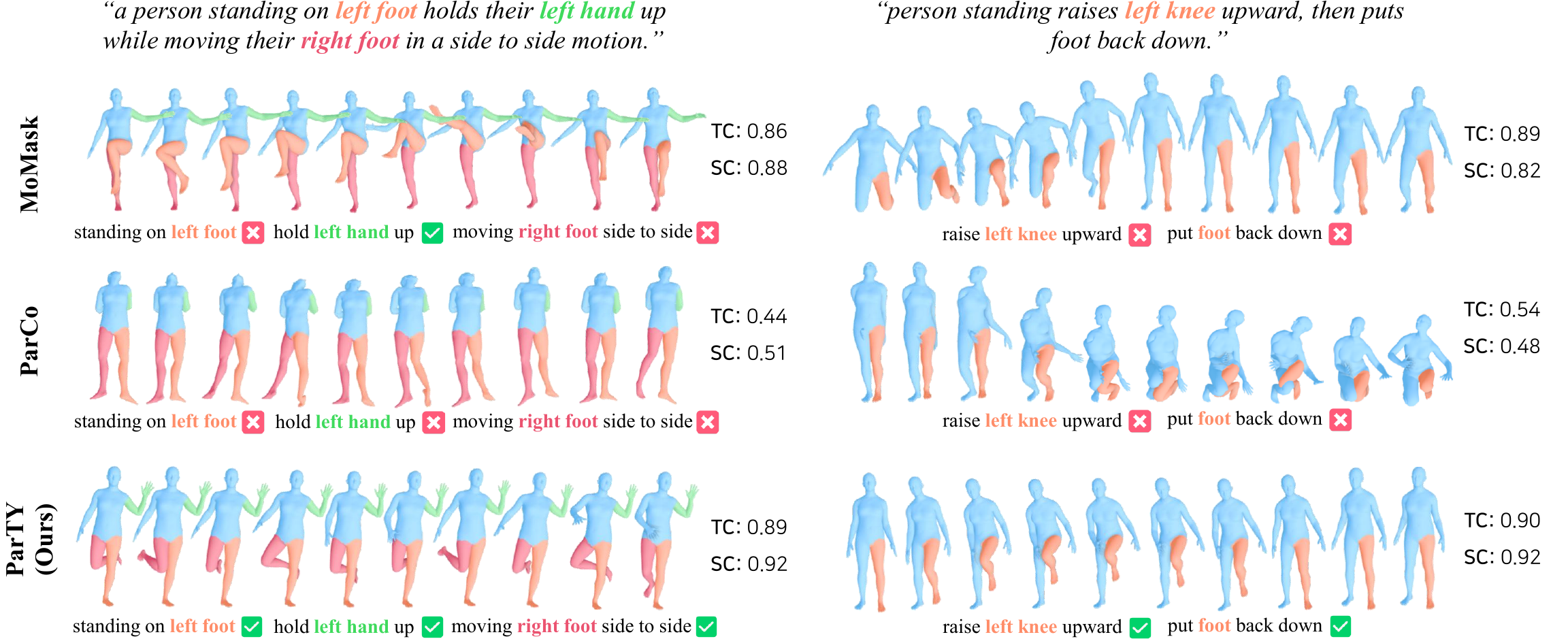}
    \caption{Qualitative comparison on HumanML3D. Colored text in the descriptions corresponds to the colored body parts in the generated motions, with coherence-level (TC, SC) scores displayed for each sample.}
    \vspace{-2mm}
    \label{fig:figure4}
\end{figure*}
\begin{equation}
S_{\text{spatial}} = \frac{1}{T} \sum_{t=1}^T \left[\sum_{(g,h) \in \mathcal{P}} s_{g,h}^{(d)}(t) + \sum_{g \in \mathcal{G}} s_g^{(\theta)}(t)\right]
\end{equation}
where $T$ is the total number of frames, and $\mathcal{G}$ is the set of body parts, averaging consistency scores across all frames and spatial relationships. TC and SC provide a comprehensive assessment of motion quality by capturing both temporal coordination and spatial plausibility of generated human motions.
\begin{table}[t]
\caption{Quantitative comparison with \textbf{part-level evaluation} metrics on HumanML3D.}
\vspace{-2mm}
\centering
\resizebox{\columnwidth}{!}{
\renewcommand{\arraystretch}{1.3}
\begin{tabular}{@{}llcccc@{}}
\toprule
Method & Part & \makecell{R-Precision\\(Top-1)$\uparrow$} & \makecell{R-Precision\\(Top-3)$\uparrow$} & FID$\downarrow$ & MM-Dist$\downarrow$ \\
\midrule
\multirow{2}{*}{MoMask~\cite{guo2024momask}} & Arms & $0.452^{\pm.003}$ & $0.761^{\pm.002}$ & $0.175^{\pm.003}$ & $3.440^{\pm.006}$ \\
& Legs & $0.403^{\pm.003}$ & $0.687^{\pm.003}$ & $0.104^{\pm.003}$ & $3.513^{\pm.009}$ \\
\midrule
\multirow{2}{*}{ParCo~\cite{zou2024parco}} & Arms & $0.468^{\pm.003}$ & $0.767^{\pm.003}$ & $0.215^{\pm.003}$ & $3.326^{\pm.008}$ \\
& Legs & $0.407^{\pm.003}$ & $0.699^{\pm.002}$ & $0.118^{\pm.003}$ & $3.482^{\pm.011}$ \\
\midrule
\multirow{2}{*}{\textbf{Ours}} & Arms & $\mathbf{0.506}^{\pm.003}$ & $\mathbf{0.802}^{\pm.002}$ & $\mathbf{0.133}^{\pm.002}$ & $\mathbf{3.079}^{\pm.005}$ \\
& Legs & $\mathbf{0.463}^{\pm.003}$ & $\mathbf{0.755}^{\pm.003}$ & $\mathbf{0.078}^{\pm.003}$ & $\mathbf{3.122}^{\pm.008}$ \\
\bottomrule
\end{tabular}
}
\label{tab:table2}
\vspace{-2mm}
\end{table}
\subsection{Quantitative Evaluation}
As shown in Tab.~\ref{tab:table1}, ParTY achieves state-of-the-art performance in R-Precision, MM-Dist, and FID metrics across both HumanML3D and KIT-ML datasets. The superior R-Precision and MM-Dist results demonstrate more precise text-motion alignment, while the leading FID scores confirm enhanced motion quality.

For part-level evaluation, as shown in Tab.~\ref{tab:table2}, ParTY outperforms both the existing part-wise method ParCo~\cite{zou2024parco} and the holistic method MoMask~\cite{guo2024momask} across all body parts in R-Precision, MM-Dist, and FID metrics. These results demonstrate that ParTY successfully achieves expressive part-level motion generation, overcoming a fundamental challenge common to both part-wise and holistic methods.

For coherence-level evaluation, we measure both temporal and spatial coherence scores and report the results in Tab.~\ref{tab:table3}. Consistent with our problem statement, ParCo, a part-wise method, shows low scores across both coherence metrics, while MoMask, a holistic method, achieves substantially more stable scores. This validates that our proposed coherence-level metrics effectively capture the coherence deficiency in part-wise methods, which was our key observation. Notably, ParTY outperforms not only ParCo but also achieves marginally better scores than MoMask, demonstrating that our Part-Guided Network with Holistic-Part Fusion module successfully addresses the coherence issues inherent in part-wise methods.

\begin{table}[t]
\caption{Quantitative comparison with \textbf{coherence-level (TC, SC) scores} on HumanML3D. We run each evaluation 20 times and report averages with 95\% confidence intervals.}
\vspace{-2mm}
\centering
\footnotesize
\begin{tabular}{@{}lcc@{}}
\toprule
Method & Temporal Coherence (TC)$\uparrow$ & Spatial Coherence (SC)$\uparrow$ \\
\midrule
ParCo~\cite{zou2024parco} & $0.49^{\pm.062}$ & $0.59^{\pm.057}$ \\
MoMask~\cite{guo2024momask} & $0.84^{\pm.047}$ & $0.90^{\pm.044}$ \\
\textbf{Ours} & $\mathbf{0.88}^{\pm.051}$ & $\mathbf{0.92}^{\pm.041}$ \\
\bottomrule
\end{tabular}
\label{tab:table3}
\vspace{-2mm}
\end{table}

\subsection{Qualitative Evaluation}
We conduct a visual comparison of motion generation results between our ParTY, ParCo~\cite{zou2024parco}, and MoMask~\cite{guo2024momask}. As shown in Fig.~\ref{fig:figure4}, both MoMask and ParCo fail to accurately reflect part-specific descriptions, whereas our ParTY accurately captures all part descriptions. Moreover, in terms of motion coherence, the part-wise method ParCo exhibits significant issues such as neck distortion and misaligned upper and lower body orientations, resulting in low temporal and spatial coherence scores. In contrast, ParTY maintains coherent motions across all frames, achieving high temporal and spatial coherence scores. These results demonstrate that ParTY effectively resolves the existing trade-off between part-specific expressiveness and full-body coherence.
\begin{table}[t]
\caption{Porting Temporal-aware VQ-VAE to MoMask~\cite{guo2024momask}. Reconstruction evaluates VQ-VAE performance, while Generation evaluates final performance including the transformer. Mean Per Joint Position Error (MPJPE) measures positional accuracy, and Average Inference Time (AIT) is averaged over 100 samples on an RTX A5000 GPU.}
\vspace{-2mm}
\centering
\resizebox{\columnwidth}{!}{
\renewcommand{\arraystretch}{1.3}
\setlength{\tabcolsep}{8pt}
\begin{tabular}{lccccccc}
\toprule
\multirow{2}{*}{Method} & \multirow{2}{*}{\makecell{Window\\size}} & \multicolumn{3}{c}{Reconstruction} & \multicolumn{2}{c}{Generation} \\
\cmidrule(lr){3-5} \cmidrule(lr){6-7}
& & \# Params & FID$\downarrow$ & MPJPE$\downarrow$ & FID$\downarrow$ & AIT \\
\midrule
MoMask & 4 & 19.44M & 0.020 & 0.030 & 0.045 & \multirow{2}{*}{80ms} \\
\cellcolor{blue!15}\hspace{0.5em}+ Ours & \cellcolor{blue!15}4 & \cellcolor{blue!15}19.86M & \cellcolor{blue!15}0.003 {\scriptsize \textcolor{blue}{(+85\%)}} & \cellcolor{blue!15}0.011 {\scriptsize \textcolor{blue}{(+63\%)}} & \cellcolor{blue!15}0.033 {\scriptsize \textcolor{blue}{(+26\%)}} & \\
\midrule
MoMask & 8 & 10.15M & 0.042 & 0.055 & 0.094 & \multirow{2}{*}{\makecell{43ms\\{\scriptsize \textcolor{red}{(+46\%)}}}} \\
\cellcolor{blue!15}\hspace{0.5em}+ Ours & \cellcolor{blue!15}8 & \cellcolor{blue!15}10.58M & \cellcolor{blue!15}0.005 {\scriptsize \textcolor{blue}{(+88\%)}} & \cellcolor{blue!15}0.014 {\scriptsize \textcolor{blue}{(+74\%)}} & \cellcolor{blue!15}0.039 {\scriptsize \textcolor{blue}{(+58\%)}} & \\
\midrule
MoMask & 12 & 7.67M & 0.079 & 0.091 & 0.126 & \multirow{2}{*}{\makecell{29ms\\{\scriptsize \textcolor{red}{(+64\%)}}}} \\
\cellcolor{blue!15}\hspace{0.5em}+ Ours & \cellcolor{blue!15}12 & \cellcolor{blue!15}8.09M & \cellcolor{blue!15}0.011 {\scriptsize \textcolor{blue}{(+86\%)}} & \cellcolor{blue!15}0.023 {\scriptsize \textcolor{blue}{(+75\%)}} & \cellcolor{blue!15}0.042 {\scriptsize \textcolor{blue}{(+67\%)}} & \\
\bottomrule
\end{tabular}
}
\vspace{-2mm}
\label{tab:table4}
\end{table}
\subsection{Discussions}
\label{sec:discussions}
All experiments were performed on the HumanML3D~\cite{guo2022generating} dataset. More detailed experimental results and analysis are provided in the supplementary material.

\noindent \textbf{Cost Efficiency of Temporal-aware VQ-VAE.} In VQ-VAE-based approaches, the window size—which defines how many consecutive frames are encoded into a single codebook entry—critically impacts both model performance and inference efficiency. To demonstrate the cost efficiency of proposed Temporal-aware VQ-VAE, we apply it to MoMask~\cite{guo2024momask} and report parameters, reconstruction, and generation performance across different window sizes in Tab.~\ref{tab:table4}. First, when porting with the same window size of 4, the LTE and GTE modules add minimal parameters but yield notable performance improvements. For MoMask-only, increasing the window size substantially reduces both model parameters and average inference time (AIT), but at the cost of significant performance degradation due to mapping longer motion sequences to each codebook. In contrast, our Temporal-aware VQ-VAE effectively prevents this degradation by capturing and preserving critical temporal information. While our Part-Guided Network inevitably increases inference time due to separate part transformers, the Temporal-aware VQ-VAE enables larger window sizes without performance loss, compensating for the increased inference cost.

\begin{figure}[t]
    \centering
    \includegraphics[width=1\linewidth]{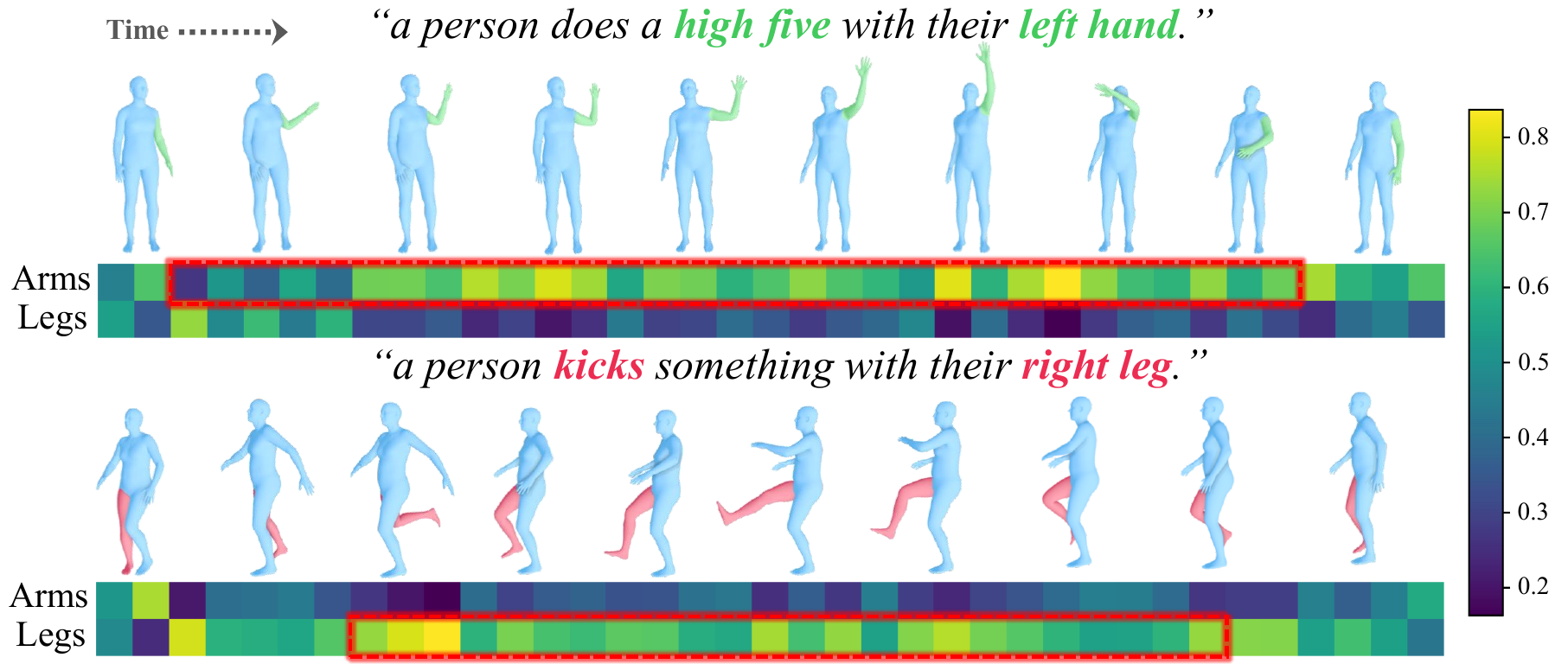}\vspace{-1mm}
    \caption{Visualization of cross attention map of HPF. Rows correspond to body parts and columns represent temporal frames. We visualize the normalized attention weights between the holistic motion token and each part motion token.}
    \vspace{-3mm}
    \label{fig:figure5}
\end{figure}

\noindent \textbf{Analysis of Part-Guided Network.} All transformers in our method autoregressively generate motion tokens, relying solely on information up to the current time step. Through Part Guidance (PG), previously generated part motion tokens are leveraged to guide holistic motion generation. The effectiveness of this approach is demonstrated in Tab.~\ref{tab:table5}, where adding PG leads to notable improvements across all metrics, showing that part information serves as valuable guidance for holistic motion generation. Furthermore, as shown in Tab.~\ref{tab:table6}, utilizing PG effectively incorporates part-level information, positively impacting part expressiveness. Additionally, Holistic-Part Fusion (HPF) enables the generation of more part-aware holistic motions by adaptively fusing holistic and part motions throughout the generation process. As shown in Fig.~\ref{fig:figure5}, for two text descriptions emphasizing different body parts, the attention scores are notably higher for the corresponding parts. This demonstrates that the HPF module effectively captures dynamic inter-part relationships and performs appropriate fusion by adaptively attending to task-relevant body parts throughout the motion sequence.

\begin{table}[t]
\caption{Ablation studies of the proposed components. PG indicates Part Guidance.}
\vspace{-2mm}
\centering
\resizebox{\columnwidth}{!}{
\renewcommand{\arraystretch}{1.3}
\begin{tabular}{@{}ccccccc@{}}
\toprule
PG & PTG & HPF & \makecell{R-Precision\\(Top-1)$\uparrow$} & \makecell{R-Precision\\(Top-3)$\uparrow$} & FID$\downarrow$ & MM-Dist$\downarrow$ \\
\midrule
& & & $0.494^{\pm.003}$ & $0.780^{\pm.003}$ & $0.158^{\pm.005}$ & $3.087^{\pm.008}$ \\
\checkmark & & & $0.520^{\pm.002}$ & $0.802^{\pm.003}$ & $0.086^{\pm.003}$ & $2.913^{\pm.010}$ \\
\checkmark & \checkmark & & $0.545^{\pm.003}$ & $0.828^{\pm.003}$ & $0.051^{\pm.003}$ & $2.799^{\pm.008}$ \\
\checkmark & \checkmark & \checkmark & $\mathbf{0.550}^{\pm.003}$ & $\mathbf{0.836}^{\pm.002}$ & $\mathbf{0.035}^{\pm.002}$ & $\mathbf{2.779}^{\pm.006}$ \\
\bottomrule
\end{tabular}
}
\label{tab:table5}
\end{table}

\begin{table}[t]
\caption{Ablation studies with \textbf{part-level evaluation} metrics.}
\vspace{-2mm}
\centering
\resizebox{\columnwidth}{!}{
\renewcommand{\arraystretch}{1.3}
\begin{tabular}{@{}lccccccc@{}}
\toprule
Part & PG & PTG & HPF & \makecell{R-Precision\\(Top-1)$\uparrow$} & \makecell{R-Precision\\(Top-3)$\uparrow$} & FID$\downarrow$ & MM-Dist$\downarrow$ \\
\midrule
\multirow{4}{*}{Arms} & & & & $0.433^{\pm.003}$ & $0.736^{\pm.002}$ & $0.232^{\pm.004}$ & $3.347^{\pm.014}$ \\
& \checkmark & & & $0.470^{\pm.003}$ & $0.769^{\pm.003}$ & $0.166^{\pm.002}$ & $3.251^{\pm.006}$ \\
& \checkmark & \checkmark & & $0.501^{\pm.003}$ & $0.798^{\pm.003}$ & $0.152^{\pm.002}$ & $3.102^{\pm.007}$ \\
& \checkmark & \checkmark & \checkmark & $\mathbf{0.506}^{\pm.003}$ & $\mathbf{0.802}^{\pm.002}$ & $\mathbf{0.133}^{\pm.002}$ & $\mathbf{3.079}^{\pm.005}$ \\
\bottomrule
\end{tabular}
}
\vspace{-2mm}
\label{tab:table6}
\end{table}

\begin{figure}[t]
    \centering
    \includegraphics[width=1\linewidth]{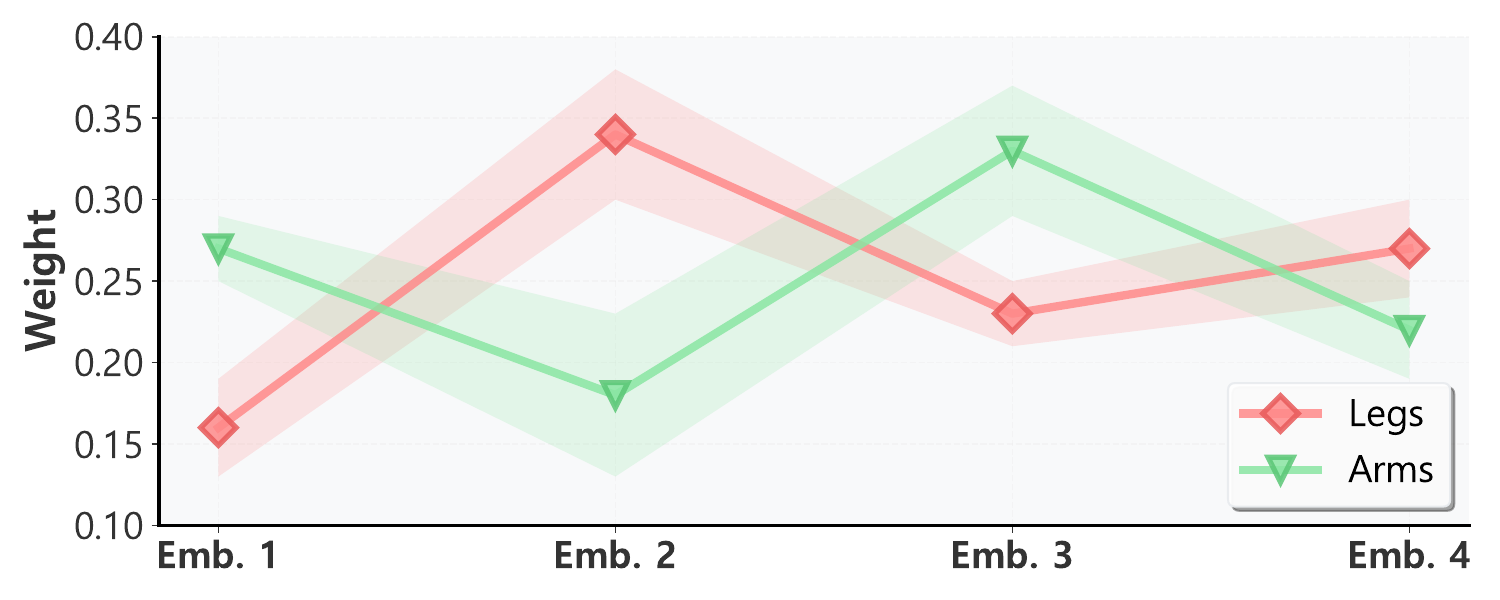}\vspace{-2mm}
    \caption{Embedding selection ratios in PTG. Mean and standard deviation of weights are computed over semantically similar text descriptions that share common motion patterns.}
    \vspace{-2mm}
    \label{fig:figure6}
\end{figure}

\noindent \textbf{Analysis of Part-aware Text Grounding.} As shown in Tab.~\ref{tab:table5} and Tab.~\ref{tab:table6}, PTG improves text-motion alignment metrics (R-Precision and MM-Dist) and part expressiveness, confirming that it enables more precise correspondence between textual descriptions and part motions. Fig.~\ref{fig:figure6} demonstrates how different embeddings are weighted and selected for specific body parts given the same input text. We adopt 4 embeddings based on experimental validation showing optimal performance. Legs show high selection ratios for embedding 2, while arms favor embedding 3, confirming that PTG effectively transforms textual embeddings into multiple specialized representations and dynamically selects the most relevant features for each body part.
%\vspace{-4mm}
\section{Conclusion}
%\vspace{-1mm}
We present ParTY, a novel model for text-driven motion generation that addresses the fundamental trade-off in existing methods. Our approach enhances part-text alignment through part-specific selection of diverse text embeddings and maintains coherent motions via a part-guided generation framework. Moreover, we introduce part-level and coherence-level evaluation metrics to comprehensively validate our model. Extensive experiments on HumanML3D and KIT-ML demonstrate that ParTY achieves state-of-the-art performance, outperforming both holistic and part-wise methods. By bridging the gap between expressive part motion and full-body motion coherence, ParTY establishes a new standard for text-to-motion generation.

\section*{Acknowledgments}
This work was supported by the National Research Foundation of Korea (NRF) grant funded by the Korea government(MSIT)(RS-2024-00456589).

{
    \small
    \bibliographystyle{ieeenat_fullname}
    \bibliography{main}
}

\newpage
\urlstyle{rm}
\def\UrlFont{\rm}
\frenchspacing
\setlength{\pdfpagewidth}{8.5in}
\setlength{\pdfpageheight}{11in}

\renewcommand{\theequation}{S\arabic{equation}}
\renewcommand{\thefigure}{S\arabic{figure}}
\renewcommand{\thetable}{S\arabic{table}}

\setcounter{equation}{0}
\setcounter{figure}{0}
\setcounter{table}{0}
\setcounter{section}{0}

\renewcommand{\thesection}{\Alph{section}}
\renewcommand{\thesubsection}{\thesection.\arabic{subsection}}
\renewcommand{\theHsection}{app.\Alph{section}}

\definecolor{mycolor}{RGB}{180, 30, 50} 

\twocolumn[{
\begin{center}
    {\Large\textbf{ParTY: Part-Guidance for Expressive Text-to-Motion Synthesis}}\\[1.2em]
    {\Large Supplementary Material}
\end{center}
\vspace{5em}
}]

\section*{Contents}

~\ref{sec:A}. Implementation Details
\begin{itemize}[leftmargin=1.5em, itemsep=1pt, label={}]
    \item ~\ref{sec:A.1}. Networks \& Training
    \item ~\ref{sec:A.2}. Body Part Division
\end{itemize}
\vspace{1mm}

\noindent ~\ref{sec:B}. Part-level Evaluation Metrics Details
\begin{itemize}[leftmargin=1.5em, itemsep=1pt, label={}]
    \item ~\ref{sec:B.1}. Design Details
    \item ~\ref{sec:B.2}. Quantitative Results \& Ablation Studies
\end{itemize}
\vspace{1mm}

\noindent ~\ref{sec:C}. Coherence-level Evaluation Metrics Details
\begin{itemize}[leftmargin=1.5em, itemsep=1pt, label={}]
    \item ~\ref{sec:C.1}. Design Details
    \item ~\ref{sec:C.2}. Hyperparameter Analysis
    \item ~\ref{sec:C.3}. Quantitative Results
    \item ~\ref{sec:C.4}. Ablation Studies
\end{itemize}
\vspace{1mm}

\noindent ~\ref{sec:D}. Additional Qualitative Results
\begin{itemize}[leftmargin=1.5em, itemsep=1pt, label={}]
    \item ~\ref{sec:D.1}. Part-Text Alignment \& Coherence
    \item ~\ref{sec:D.2}. Long \& Complex
\end{itemize}
\vspace{1mm}

\noindent ~\ref{sec:E}. Additional Quantitative Results
\begin{itemize}[leftmargin=1.5em, itemsep=1pt, label={}]
    \item ~\ref{sec:E.1}. Detailed Results \& Ablation Studies
    \item ~\ref{sec:E.2}. Computational Complexity
    \item ~\ref{sec:E.3}. Temporal-aware VQ-VAE
    \begin{itemize}[leftmargin=1.5em, itemsep=1pt, label={}]
        \item ~\ref{sec:E.3.1}. Porting to Additional Models
        \item ~\ref{sec:E.3.2}. Ablation Studies
    \end{itemize}
    \item ~\ref{sec:E.4}. Part-aware Text Grounding
    \begin{itemize}[leftmargin=1.5em, itemsep=1pt, label={}]
        \item ~\ref{sec:E.4.1}. Analysis of the Number of Embeddings
        \item ~\ref{sec:E.4.2}. Ablation Studies on Auxilary Loss
        \item ~\ref{sec:E.4.3}. LLM Prompt Details
        \item ~\ref{sec:E.4.4}. Quality Evaluation of Generated Text
        \item ~\ref{sec:E.4.5}. Group of Text Descriptions
    \end{itemize}
    \item ~\ref{sec:E.5}. Part Guidance
    \begin{itemize}[leftmargin=1.5em, itemsep=1pt, label={}]
        \item ~\ref{sec:E.5.1}. Analysis of the Size of Window
    \end{itemize}
    \item ~\ref{sec:E.6}. Holistic-Part Fusion
    \begin{itemize}[leftmargin=1.5em, itemsep=1pt, label={}]
        \item ~\ref{sec:E.6.1}. Additional Visualization of Attention Map
    \end{itemize}
\end{itemize}
\vspace{1mm}

\noindent ~\ref{sec:F}. User Study Details
\vspace{1mm}

% \noindent ~\ref{sec:G}. Limitations

\begin{figure*}[t]
    \centering
    \includegraphics[width=1\linewidth]{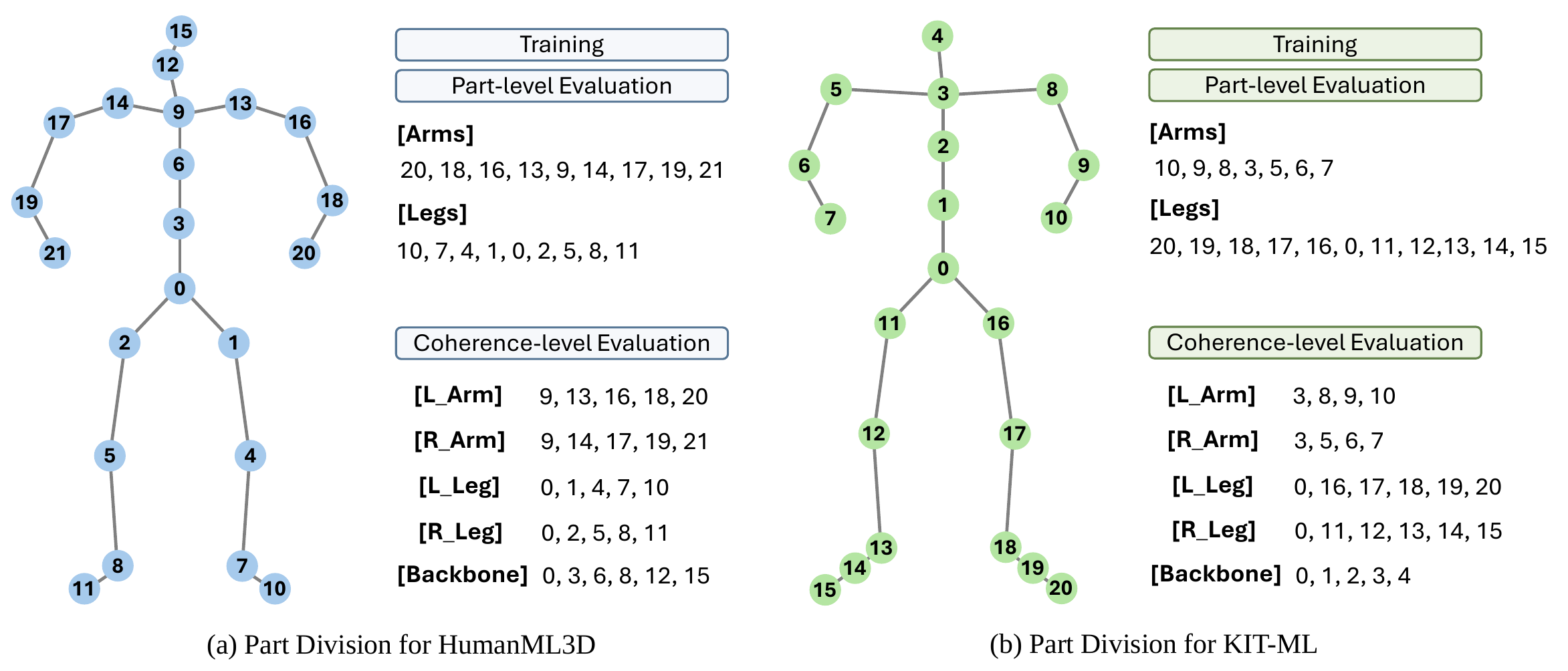}
    \caption{Body part division for HumanML3D and KIT-ML.}
    \label{fig:sup_div}
\end{figure*}

\newpage

\section{Implementation Details}\label{sec:A}
\subsection{Networks \& Training}\label{sec:A.1}
\noindent \textbf{Temporal-aware VQ-VAE} utilizes a codebook with 256 entries and a code dimension of 128. In the Local Temporal Enhancement (LTE) module, we set the window size to 8. For Global Temporal Enhancement (GTE), we adopt a 3-layer GCN, where each layer has a hidden dimension of 128. The part VQ-VAE uses a codebook with 128 entries and the same code dimension of 128. In its LTE module, the window size is set to 12, while the GTE configuration remains identical to that of the holistic VQ-VAE. For training, we employ the AdamW~\cite{loshchilov2017decoupled} optimizer with $\beta_1 = 0.9$ and $\beta_2 = 0.99$. The learning rate follows a two-stage schedule: $2 \times 10^{-4}$ for the first 200K iterations, and $1 \times 10^{-5}$ for the remaining iterations. We use a batch size of 256 and set the approximation loss weight to $\mathcal{L}_{app} = 1.0$.

\noindent \textbf{Part-aware Text Grounding} module employs 4 MLPs (3-layer with ReLU activation) to diversely transform the text embeddings. This configuration was determined to yield optimal performance through extensive experimental validation, as detailed in Sec.~\ref{sec:E.4.1}. For the contrastive learning objective, we set the temperature parameter to $\tau = 0.05$. In the overall loss function, the diversity loss $\mathcal{L}_{div}$ is weighted by $\lambda_{div} = 0.1$. The gating network for the Part Gate is implemented as a simple linear layer. For LLM-generated part text supervision, we apply the auxiliary loss $\mathcal{L}_{\text{aux}}$ with a weighting factor of $\lambda_{aux} = 0.1$.

\noindent \textbf{Part-Guided Network} constructs Part Guidance from three time steps of the part transformer, a design choice chosen based on extensive experimental validation (Sec.~\ref{sec:E.5.1}), and employs a holistic transformer with 10 layers and a token dimension of 128, as well as a part transformer with 8 layers and the same token dimension of 128. In the \textbf{Holistic-Part Fusion} module, both the self-attention and cross-attention layers use 6 attention heads with a head dimension of 64. For training, we also use the AdamW optimizer with $\beta_1 = 0.5$ and $\beta_2 = 0.99$. The learning rate is set to $2 \times 10^{-4}$ for the initial 100K iterations and then reduced to $5 \times 10^{-6}$ for the remaining iterations. We use a batch size of 64 to accommodate the memory requirements of the transformer architecture, and all experiments are conducted on a single NVIDIA A5000 GPU.

\subsection{Body Part Division}\label{sec:A.2}
Fig.~\ref{fig:sup_div} illustrates how we construct the Legs and Arms for training and part-level evaluation on both the HumanML3D~\cite{guo2022generating} and KIT-ML~\cite{plappert2016kit} datasets. It also shows how the five parts (Left Arm, Right Arm, Left Leg, Right Leg, Backbone) are defined for coherence-level evaluation.

\section{Part-level Evaluation Metrics Details}\label{sec:B}
\subsection{Design Details}\label{sec:B.1}
T2M~\cite{guo2022generating} proposes a motion encoder for evaluating generated motions; however, this encoder is designed for assessing holistic motion. To enable part-level evaluation, we train this motion encoder separately on arms and legs data, resulting in specialized part motion encoders capable of evaluating motion quality at the part level. R-Precision, FID, and MM-Dist are computed following the same evaluation protocol as in T2M~\cite{guo2022generating}.

\subsection{Quantitative Results \& Ablation Studies}\label{sec:B.2}
We demonstrate the generality of our metrics by reporting part-level evaluation results on additional models~\cite{zhong2023attt2m, sun2024lgtm} in Tab.~\ref{tab:tables1}. Furthermore, Tab.~\ref{tab:tables2} reports the part-level evaluation results for the legs in our proposed methods.

\begin{table}[t]
\caption{Quantitative comparison with \textbf{part-level evaluation} metrics on HumanML3D.}
\centering
\resizebox{\columnwidth}{!}{
\renewcommand{\arraystretch}{1.3}
\begin{tabular}{@{}llcccc@{}}
\toprule
Method & Part & \makecell{R-Precision\\(Top-1)$\uparrow$} & \makecell{R-Precision\\(Top-3)$\uparrow$} & FID$\downarrow$ & MM-Dist$\downarrow$ \\
\midrule
\multirow{2}{*}{MoMask~\cite{guo2024momask}} & Arms & $0.452^{\pm.003}$ & $0.761^{\pm.002}$ & $0.175^{\pm.003}$ & $3.440^{\pm.006}$ \\
& Legs & $0.403^{\pm.003}$ & $0.687^{\pm.003}$ & $0.104^{\pm.003}$ & $3.513^{\pm.009}$ \\
\midrule
\multirow{2}{*}{AttT2M~\cite{zhong2023attt2m}} & Arms & $0.431^{\pm.003}$ & $0.738^{\pm.002}$ & $0.264^{\pm.002}$ & $3.489^{\pm.010}$ \\
& Legs & $0.375^{\pm.003}$ & $0.656^{\pm.003}$ & $0.190^{\pm.003}$ & $3.541^{\pm.011}$ \\
\midrule
\multirow{2}{*}{ParCo~\cite{zou2024parco}} & Arms & $0.468^{\pm.003}$ & $0.767^{\pm.003}$ & $0.215^{\pm.003}$ & $3.326^{\pm.008}$ \\
& Legs & $0.407^{\pm.003}$ & $0.699^{\pm.002}$ & $0.118^{\pm.003}$ & $3.482^{\pm.011}$ \\
\midrule
\multirow{2}{*}{LGTM~\cite{sun2024lgtm}} & Arms & $0.436^{\pm.003}$ & $0.745^{\pm.003}$ & $0.398^{\pm.002}$ & $3.421^{\pm.015}$ \\
& Legs & $0.384^{\pm.003}$ & $0.689^{\pm.003}$ & $0.325^{\pm.002}$ & $3.547^{\pm.018}$ \\
\midrule
\multirow{2}{*}{\textbf{Ours}} & Arms & $\mathbf{0.506}^{\pm.003}$ & $\mathbf{0.802}^{\pm.002}$ & $\mathbf{0.133}^{\pm.002}$ & $\mathbf{3.079}^{\pm.005}$ \\
& Legs & $\mathbf{0.463}^{\pm.003}$ & $\mathbf{0.755}^{\pm.003}$ & $\mathbf{0.078}^{\pm.003}$ & $\mathbf{3.122}^{\pm.008}$ \\
\bottomrule
\end{tabular}
}
\label{tab:tables1}
\end{table}

\begin{table}[t]
\caption{Ablation studies with \textbf{part-level evaluation} metrics on HumanML3D.}
\centering
\resizebox{\columnwidth}{!}{
\renewcommand{\arraystretch}{1.3}
\begin{tabular}{@{}lccccccc@{}}
\toprule
Part & PG & PTG & HPF & \makecell{R-Precision\\(Top-1)$\uparrow$} & \makecell{R-Precision\\(Top-3)$\uparrow$} & FID$\downarrow$ & MM-Dist$\downarrow$ \\
\midrule
\multirow{4}{*}{Legs} & & & & $0.397^{\pm.003}$ & $0.691^{\pm.003}$ & $0.169^{\pm.003}$ & $3.416^{\pm.012}$ \\
& \checkmark & & & $0.422^{\pm.003}$ & $0.715^{\pm.003}$ & $0.114^{\pm.003}$ & $3.328^{\pm.011}$ \\
& \checkmark & \checkmark & & $0.456^{\pm.002}$ & $0.744^{\pm.003}$ & $0.095^{\pm.003}$ & $3.175^{\pm.006}$ \\
& \checkmark & \checkmark & \checkmark & $\mathbf{0.463}^{\pm.003}$ & $\mathbf{0.755}^{\pm.003}$ & $\mathbf{0.078}^{\pm.003}$ & $\mathbf{3.122}^{\pm.008}$ \\
\bottomrule
\end{tabular}
}
\label{tab:tables2}
\end{table}

\section{Coherence-level Evaluation Metrics Details}\label{sec:C} % 수정 예정
\subsection{Design Details}\label{sec:C.1}
To evaluate motion coherence at the frame-level, we introduce the Temporal Coherence (TC) and Spatial Coherence (SC) score, which evaluate both temporal and spatial consistency across body parts. A motion sequence is represented by $j$ joints with 3D position $\hat{\mathbf{p}}_j(t)$ at time step $t$, partitioned into five body parts: left arm, right arm, left leg, right leg, and backbone.

\noindent \textbf{Temporal Coherence} score quantifies temporal coordination between  body parts over time. To quantify the instantaneous motion strength of each part in a noise-robust manner, we compute the temporal-wise root mean square (RMS) velocity for each body part $g$, defined as:

\begin{equation}
\mathbf{x}_g(t) = {\sqrt{\frac{1}{n_g} \sum_{j \in g} \|\hat{\mathbf{p}}_j(t) - \hat{\mathbf{p}}_j(t-1)\|^2}}
\end{equation}

\noindent where the sum is over all joints $j$ belonging to part $g$, and $n_g$ is the number of joints from part $g$. To compare motion patterns across parts with different movement intensities, we apply z-normalization within sliding windows $w$ of length $L$ frames with stride $L/2$ frames—local time intervals that allow adaptive normalization to account for varying motion dynamics throughout the sequence. This standardization (zero mean, unit variance) ensures that the subsequent cross-correlation reflects only the relative phase and shape similarity of motion signals rather than their absolute magnitudes:

\begin{equation}
\small
\begin{gathered}
\bar{x}_g^{(w)} = \frac{1}{|\mathcal{T}_w|} \sum_{t \in \mathcal{T}_w} x_g(t), \quad \sigma_g^{(w)} = \sqrt{\frac{1}{|\mathcal{T}_w|} \sum_{t \in \mathcal{T}_w} (x_g(t) - \bar{x}_g^{(w)})^2} \\
s_g^{(w)}(t) = \frac{x_g(t) - \bar{x}_g^{(w)}}{\sigma_g^{(w)} + \varepsilon}, \quad t \in \mathcal{T}_w
\end{gathered}
\end{equation}

\noindent where $\mathcal{T}_w$ is the set of frame indices in the $w$-th sliding window, $\bar{x}_g^{(w)}$ and $\sigma_g^{(w)}$ denote the local mean and standard deviation of the part-wise RMS velocity, and $\varepsilon$ is a small constant for numerical stability. For each part pair $(g,h)$, we compute the pairwise cross-correlation:

\begin{equation}
r_{g,h}^{(w)}(\tau) = \frac{\sum_{t \in \mathcal{T}_w} s_g^{(w)}(t) s_h^{(w)}(t+\tau)}{\sqrt{\sum_{t \in \mathcal{T}_w} (s_g^{(w)}(t))^2} \sqrt{\sum_{t \in \mathcal{T}_w} (s_h^{(w)}(t))^2}}
\end{equation}

\noindent where $\tau \in [-\tau_{\max}, \tau_{\max}]$ represents the temporal lag (time shift in frames) between the motion signals of parts $g$ and $h$. By computing cross-correlation across multiple lags, we can detect phase-shifted synchrony—coordinated movements even when body parts move with natural timing offsets—for instance, in walking, arm motion naturally leads or lags leg motion. We aggregate correlation values across all lags using a softmax-weighted average, retain only positive correlations (in-phase movements), and apply an exponential penalty based on the expected lag magnitude to suppress spurious correlations from unrelated movements or excessive delays:

\begin{equation}
R_{g,h}^{(w)} = \mathbb{E}_{\tau}\left[r_{g,h}^{(w)}(\tau)\right], \quad \langle |\tau| \rangle_w = \mathbb{E}_{\tau}\left[|\tau|\right]
\end{equation}
\begin{equation}
\tilde{R}_{g,h}^{(w)} = \max(0, R_{g,h}^{(w)}) \cdot \exp\left(-\frac{\langle |\tau| \rangle_w}{\kappa}\right)
\end{equation}

\noindent where $\mathbb{E}_{\tau}[\cdot] = \frac{\sum_{\tau} (\cdot) \exp(r_{g,h}^{(w)}(\tau)/\sigma)}{\sum_{\tau} \exp(r_{g,h}^{(w)}(\tau)/\sigma)}$ denotes the softmax-weighted expectation over temporal lags with temperature $\sigma$, $\langle |\tau| \rangle_w$ measures the expected absolute lag (average timing offset between coordinated movements), and $\kappa$ controls the strength of the delay penalty. The temporal coherence score is then computed as:

\begin{equation}
S_{\text{temporal}} = \frac{1}{W} \sum_{w=1}^W \frac{1}{|\mathcal{P}|} \sum_{(g,h) \in \mathcal{P}} \tilde{R}_{g,h}^{(w)}
\end{equation}

\noindent where $W$ is the total number of sliding windows, $\mathcal{P}$ denotes the set of all body part pairs, yielding a measure of overall rhythmic coordination.

\noindent \textbf{Spatial Coherence} score evaluates the physical plausibility of spatial relationships within each frame. To capture the overall spatial configuration of each part while being robust to joint-level noise, we define the representative position of part $g$ as the average position of its joints:

\begin{equation}
\mathbf{c}_g(t) = \frac{1}{n_g} \sum_{j \in g} \hat{\mathbf{p}}_j(t)
\end{equation}

\noindent Using these representative points, we measure inter-part distances $d_{g,h}(t) = \|\mathbf{c}_g(t) - \mathbf{c}_h(t)\|$ to assess global spatial relationships and angles relative to the torso to evaluate local anatomical plausibility and articulation consistency:

\begin{equation}
\mathbf{u}_g(t) = \frac{\Delta \hat{\mathbf{p}}_g(t)}{\|\Delta \hat{\mathbf{p}}_g(t)\|}, \quad \theta_g(t) = \arccos(\mathbf{u}_g(t) \cdot \mathbf{u}_{\text{TR}}(t))
\end{equation}
\noindent where $\Delta \hat{\mathbf{p}}_g(t) = \hat{\mathbf{p}}_{\text{end}}(t) - \mathbf{c}_g(t)$ is the part direction vector from $\mathbf{c}_g(t)$ to the end joint of part $g$, $\mathbf{u}_g(t)$ and $\mathbf{u}_{\text{TR}}(t)$ are the normalized part and torso orientation vectors, and $\theta_g(t)$ is the articulation angle.

Using statistics computed from the HumanML3D~\cite{guo2022generating}, we normalize the distance and angle measurements to obtain z-scores that quantify how much each frame deviates from typical human motion patterns. The normalized scores are then converted to spatial consistency scores using Gaussian kernels:

\begin{equation}
\small
\begin{aligned}
z_{g,h}^{(d)}(t) &= \frac{d_{g,h}(t) - \mu_{g,h}^{(d)}}{\sigma_{g,h}^{(d)} + \varepsilon}, \quad z_g^{(\theta)}(t) = \frac{\theta_g(t) - \mu_g^{(\theta)}}{\sigma_g^{(\theta)} + \varepsilon} \\
s_{g,h}^{(d)}(t) &= \exp\left(-\frac{(z_{g,h}^{(d)}(t))^2}{\beta_d^2}\right), \quad s_g^{(\theta)}(t) = \exp\left(-\frac{(z_g^{(\theta)}(t))^2}{\beta_\theta^2}\right)
\end{aligned}
\end{equation}

\noindent where $\mu_{g,h}^{(d)}$ and $\sigma_{g,h}^{(d)}$ are the mean and standard deviation of inter-part distance $d_{g,h}$ in HumanML3D, $\mu_g^{(\theta)}$ and $\sigma_g^{(\theta)}$ are the statistics for articulation angle $\theta_g$, $\varepsilon$ is a small constant for numerical stability, and $\beta_d$ and $\beta_\theta$ are bandwidth parameters controlling the sensitivity to deviations. The spatial coherence score is then computed by averaging these consistency scores across all frames:

\begin{table}[t]
\small
\centering
\caption{Ablation studies on $\sigma$ and $\kappa$ parameters.}
\label{tab:ablation}
\begin{tabular}{lc}
\toprule
Hyperparameters & Temporal Coherence $\uparrow$ \\
\midrule
$\sigma=0.05$, $\kappa=3$ & 0.92 \\
$\sigma=0.05$, $\kappa=5$ & 0.93 \\
$\sigma=0.05$, $\kappa=10$ & 0.93 \\
$\sigma=0.05$, $\kappa=15$ & 0.91 \\
\midrule
$\sigma=0.1$, $\kappa=3$ & 0.95 \\

\rowcolor{RoyalBlue!20}
$\bm{\sigma}=\mathbf{0.1}$, $\bm{\kappa}=\mathbf{5}$ & $\mathbf{0.96}$ \\

$\sigma=0.1$, $\kappa=10$ & 0.94 \\
$\sigma=0.1$, $\kappa=15$ & 0.93 \\
\midrule
$\sigma=0.2$, $\kappa=3$ & 0.93 \\
$\sigma=0.2$, $\kappa=5$ & 0.94 \\
$\sigma=0.2$, $\kappa=10$ & 0.92 \\
$\sigma=0.2$, $\kappa=15$ & 0.91 \\
\midrule
$\sigma=0.5$, $\kappa=3$ & 0.89 \\
$\sigma=0.5$, $\kappa=5$ & 0.91 \\
$\sigma=0.5$, $\kappa=10$ & 0.88 \\
$\sigma=0.5$, $\kappa=15$ & 0.86 \\
\bottomrule
\end{tabular}
\label{tab:tables3}
\end{table}

\begin{table}[t]
\small
\centering
\caption{Ablation studies on $L$ and $\tau_{\text{max}}$ parameters.}
\label{tab:tables4}
\begin{tabular}{lc}
\toprule
Method & Temporal Coherence $\uparrow$ \\
\midrule
$L=20$, $\tau_{\text{max}}=10$ & 0.95 \\

\rowcolor{RoyalBlue!20}
$\bm{L}=\mathbf{20}$, $\bm{\tau}_{\text{\textbf{max}}}=\mathbf{15}$ & $\mathbf{0.96}$ \\

\rowcolor{RoyalBlue!20}
$\bm{L}=\mathbf{20}$, $\bm{\tau}_{\text{\textbf{max}}}=\mathbf{20}$ & $\mathbf{0.96}$ \\

$L=20$, $\tau_{\text{max}}=30$ & 0.94 \\
\midrule
$L=30$, $\tau_{\text{max}}=10$ & 0.94 \\
$L=30$, $\tau_{\text{max}}=15$ & 0.95 \\
$L=30$, $\tau_{\text{max}}=20$ & 0.93 \\
$L=30$, $\tau_{\text{max}}=30$ & 0.90 \\
\midrule
$L=40$, $\tau_{\text{max}}=10$ & 0.89 \\
$L=40$, $\tau_{\text{max}}=15$ & 0.89 \\
$L=40$, $\tau_{\text{max}}=20$ & 0.86 \\
$L=40$, $\tau_{\text{max}}=30$ & 0.82 \\
\bottomrule
\end{tabular}
\label{tab:tables4}
\end{table}

\begin{table}[t]
\small
\centering
\caption{Ablation studies on $\beta_d$ and $\beta_\theta$ parameters.}
\label{tab:ablation_beta}
\begin{tabular}{ccc}
\toprule
$\beta_d$ & $\beta_\theta$ & Spatial Coherence $\uparrow$ \\
\midrule
1.0 & 1.0 & 0.97 \\

\rowcolor{RoyalBlue!20}
$\mathbf{1.5}$ & $\mathbf{1.5}$ & $\mathbf{0.99}$ \\

2.0 & 2.0 & 0.98 \\
3.0 & 3.0 & 0.94 \\
\bottomrule
\end{tabular}
\label{tab:tables5}
\end{table}

\begin{equation}
S_{\text{spatial}} = \frac{1}{T} \sum_{t=1}^T \left[\sum_{(g,h) \in \mathcal{P}} s_{g,h}^{(d)}(t) + \sum_{g \in \mathcal{G}} s_g^{(\theta)}(t)\right]
\end{equation}
\noindent where $T$ is the total number of frames, and $\mathcal{G}$ represents the set of all body parts.

\subsection{Hyperparameter Analysis}\label{sec:C.2}
We experimented with various combinations of hyperparameters in our coherence-level evaluation metrics (TC, SC) to find optimal values against ground truth motions from HumanML3D. Results for Temporal Coherence are reported in Tab.~\ref{tab:tables3} and Tab.~\ref{tab:tables4}, while results for Spatial Coherence are reported in Tab.~\ref{tab:tables5}.

\subsection{Quantitative Results}\label{sec:C.3}
We demonstrate the generality of our metrics by reporting coherence-level evaluation results on additional models~\cite{sun2024lgtm, zhang2023generating} in Tab.~\ref{tab:tables6}. Tab.~\ref{tab:tables7} presents the results on an additional dataset, KIT-ML~\cite{plappert2016kit}, following the same evaluation process as HumanML3D~\cite{guo2022generating}. The optimal hyperparameters for KIT-ML are $(\sigma=0.05,\ \kappa=5)$, $(L=20,\ \tau_{\text{max}}=15)$, and $(\beta_d=\beta_\theta=1.5)$.

\begin{table}[t]
\caption{Quantitative comparison with \textbf{coherence-level (TC, SC) scores} on HumanML3D.}
\centering
\footnotesize
\begin{tabular}{@{}lcc@{}}
\toprule
Method & Temporal Coherence (TC)$\uparrow$ & Spatial Coherence (SC)$\uparrow$ \\
\midrule
Real motion & $0.96^{\pm.032}$ & $0.99^{\pm.026}$ \\
\midrule
ParCo~\cite{zou2024parco} & $0.49^{\pm.062}$ & $0.59^{\pm.057}$ \\
LGTM~\cite{sun2024lgtm} & $0.51^{\pm.048}$ & $0.65^{\pm.039}$ \\
T2M-GPT~\cite{zhang2023generating} & $0.85^{\pm.050}$ & $0.87^{\pm.036}$ \\
MoMask~\cite{guo2024momask} & $0.84^{\pm.047}$ & $0.90^{\pm.044}$ \\
\textbf{Ours} & $\mathbf{0.88}^{\pm.051}$ & $\mathbf{0.92}^{\pm.041}$ \\
\bottomrule
\end{tabular}
\label{tab:tables6}
\end{table}

\begin{table}[t]
\caption{Quantitative comparison with \textbf{coherence-level (TC, SC) scores} on KIT-ML.}
\centering
\footnotesize
\begin{tabular}{@{}lcc@{}}
\toprule
Method & Temporal Coherence (TC)$\uparrow$ & Spatial Coherence (SC)$\uparrow$ \\
\midrule
Real motion & $0.87^{\pm.045}$ & $0.91^{\pm.037}$ \\
\midrule
ParCo~\cite{zou2024parco} & $0.32^{\pm.066}$ & $0.44^{\pm.062}$ \\
LGTM~\cite{sun2024lgtm} & $0.40^{\pm.061}$ & $0.49^{\pm.058}$ \\
T2M-GPT~\cite{zhang2023generating} & $0.77^{\pm.057}$ & $0.80^{\pm.049}$ \\
MoMask~\cite{guo2024momask} & $0.79^{\pm.042}$ & $0.79^{\pm.065}$ \\
\textbf{Ours} & $\mathbf{0.80}^{\pm.048}$ & $\mathbf{0.81}^{\pm.054}$ \\
\bottomrule
\end{tabular}
\label{tab:tables7}
\end{table}

\begin{table}[t]
\footnotesize
\centering
\caption{Ablation studies with \textbf{coherence-level evaluation} metrics on HumanML3D.}
\label{tab:ablation_components}
\begin{tabular}{ccccc}
\toprule
PG & PTG & HPF & Temporal Coherence$\uparrow$ & Spatial Coherence$\uparrow$ \\
\midrule
   &     &     & $0.43^{\pm.056}$ & $0.54^{\pm.044}$ \\
\checkmark &     &     & $0.81^{\pm.048}$ & $0.86^{\pm.029}$ \\
\checkmark & \checkmark &     & $0.83^{\pm.045}$ & $0.89^{\pm.034}$ \\
\checkmark & \checkmark & \checkmark & $\mathbf{0.88}^{\pm.051}$ & $\mathbf{0.92}^{\pm.041}$ \\
\bottomrule
\vspace{-10mm}
\end{tabular}
\label{tab:tables8}
\end{table}

\begin{figure*}[t]
    \centering
    \includegraphics[width=1\linewidth]{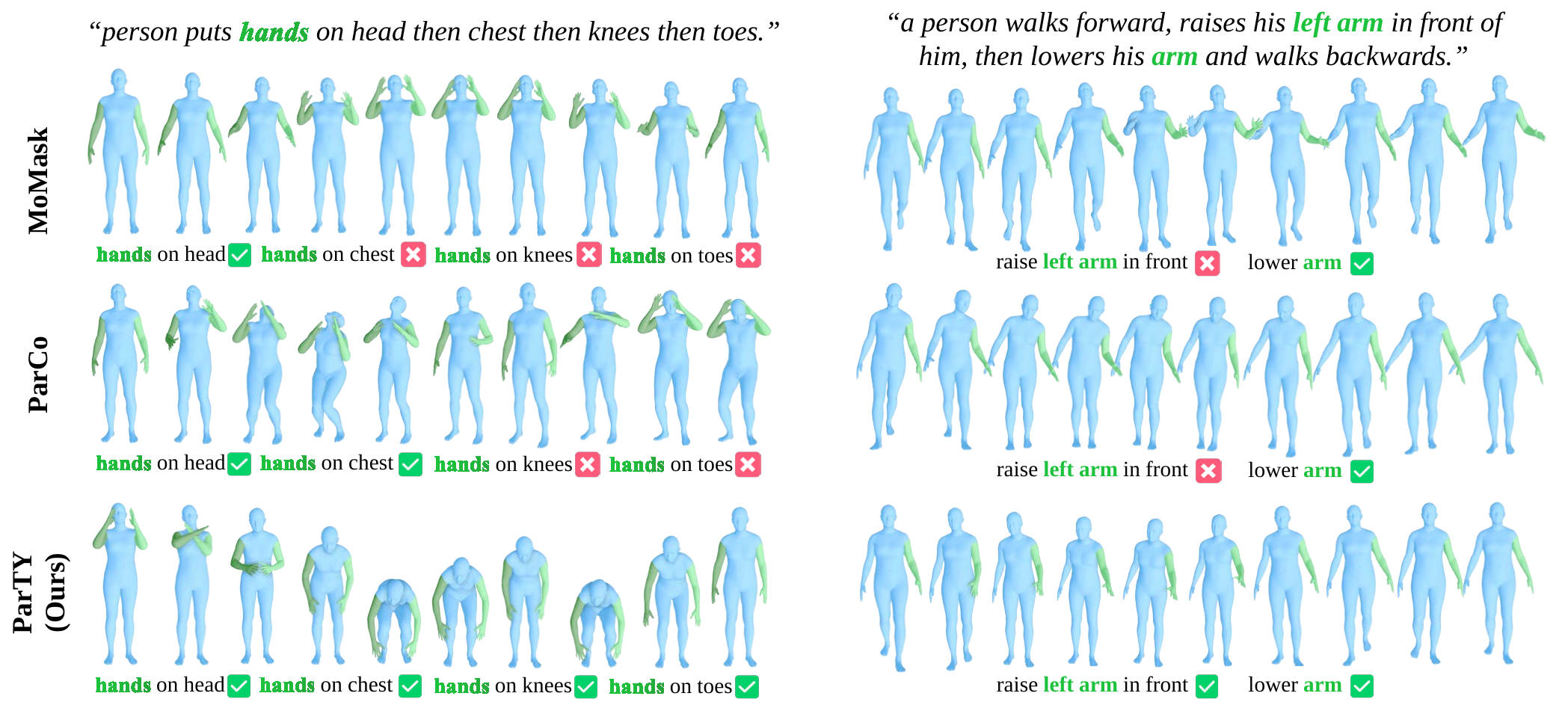}
    \caption{Qualitative comparison on part-text alignment on HumanML3D.}
    \label{fig:sup_vis_1}
\end{figure*}

\begin{figure*}[t]
    \centering
    \includegraphics[width=1\linewidth]{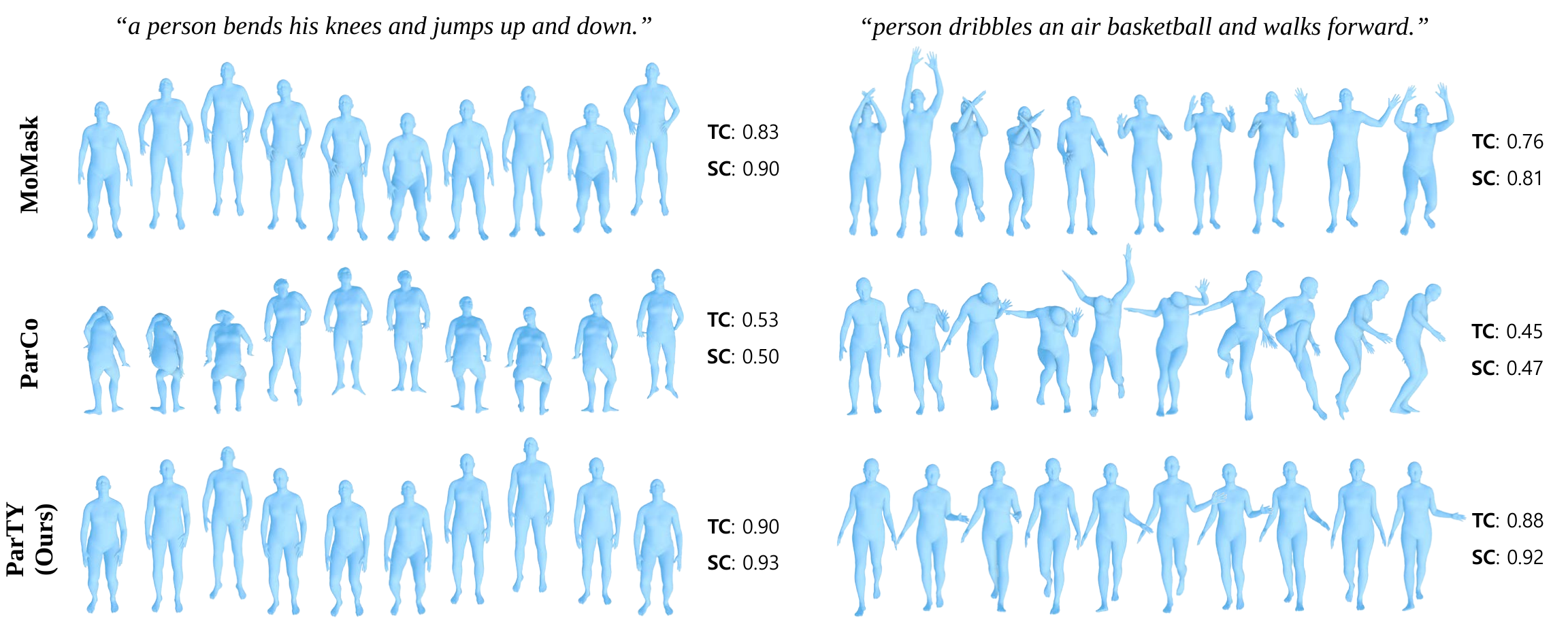}
    \caption{Qualitative comparison on coherence on HumanML3D.}
    \label{fig:sup_vis_2}
\end{figure*}

\subsection{Ablation Studies}\label{sec:C.4}
We provide ablation study results for the proposed components on coherence-level evaluation in Tab.~\ref{tab:tables8}. The results demonstrate that our ParTY, which leverages a Part-Guided Network, successfully preserves coherence while enhancing part expressiveness, compared to conventional part-wise approaches that independently generate part motions and simply integrate them.

\section{Additional Qualitative Results}\label{sec:D}

\begin{figure*}[t]
    \centering
    \includegraphics[width=1\linewidth]{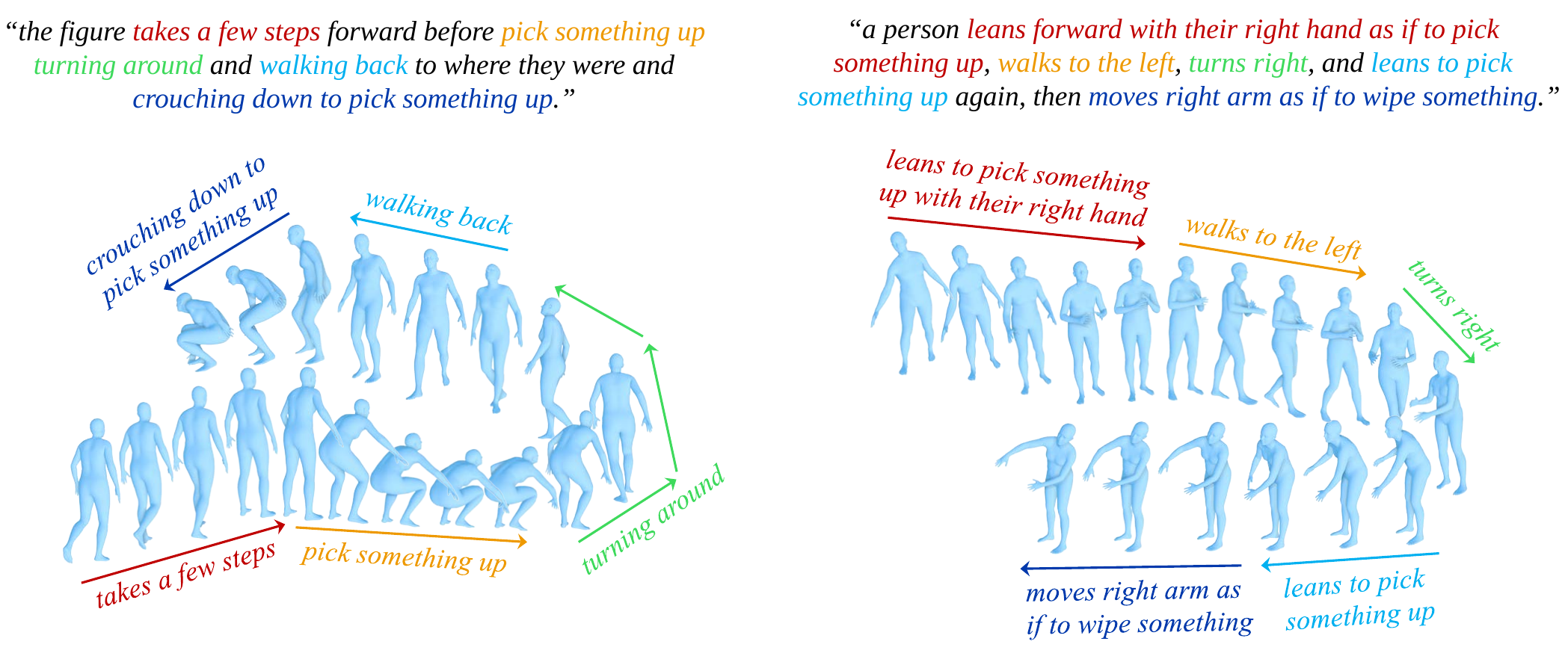}
    \caption{Qualitative comparison on long and complex motions on HumanML3D.}
    \label{fig:sup_vis_3}
\end{figure*}

\subsection{Part-Text Alignment \& Coherence}\label{sec:D.1}
Fig.~\ref{fig:sup_vis_1} and Fig.~\ref{fig:sup_vis_2} show additional samples for part-text alignment and coherence, respectively. ParTY demonstrates strong performance in both aspects.

\subsection{Long \& Complex}\label{sec:D.2}
Fig.~\ref{fig:sup_vis_3} shows that ParTY accurately handles each action throughout long and complex motions (over 170 frames).

\begin{table*}[t]
  \caption{Comprehensive ablation studies of the proposed component. \textbf{Bold} indicates the best result, while \underline{underlined} refers the second-best. The right arrow $\rightarrow$ indicates that closer values to ground truth are preferred.}
  \centering
  \small
  \begin{tabular}{cccccccccc}
    \toprule
    \multirow{2}{*}{PG} & \multirow{2}{*}{PTG} & \multirow{2}{*}{HPF}
    & \multicolumn{3}{c}{R-Precision $\uparrow$} 
    & \multirow{2}{*}{FID $\downarrow$} 
    & \multirow{2}{*}{MM-Dist $\downarrow$} 
    & \multirow{2}{*}{Diversity $\rightarrow$} 
    & \multirow{2}{*}{MModality $\uparrow$} \\
    \cmidrule(lr){4-6}
    & & & Top-1 & Top-2 & Top-3 & & & & \\
    \midrule

    & & &
    $0.494^{\pm .003}$ & $0.686^{\pm .003}$ & $0.780^{\pm .003}$ &
    $0.158^{\pm .005}$ & $3.087^{\pm .008}$ & $9.722^{\pm .067}$ & $1.445^{\pm .051}$ \\

    \checkmark & & &
    $0.520^{\pm.002}$ &
    $0.711^{\pm.003}$ &
    $0.802^{\pm.003}$ &
    $0.086^{\pm.003}$ &
    $2.913^{\pm.010}$ & $9.688^{\pm.045}$ & $1.841^{\pm.079}$ \\

    & & \checkmark &
    $0.506^{\pm.003}$ &
    $0.695^{\pm.003}$ &
    $0.791^{\pm.003}$ &
    $0.132^{\pm.003}$ &
    $3.019^{\pm.013}$ & $9.710^{\pm.056}$ & $1.860^{\pm.064}$ \\

    \checkmark & \checkmark & &
    $\underline{0.545}^{\pm.003}$ &
    $\underline{0.734}^{\pm.003}$ &
    $\underline{0.828}^{\pm.003}$ &
    $\underline{0.051}^{\pm.003}$ &
    $\underline{2.799}^{\pm.010}$ & $9.556^{\pm.042}$ & $1.973^{\pm.068}$ \\

    \checkmark & & \checkmark &
    $0.529^{\pm.002}$ &
    $0.721^{\pm.003}$ &
    $0.813^{\pm.003}$ &
    $0.063^{\pm.002}$ &
    $2.858^{\pm.008}$ & $\underline{9.528}^{\pm.059}$ & $\underline{2.101}^{\pm.071}$ \\

    & \checkmark & \checkmark &
    $0.538^{\pm.003}$ &
    $0.729^{\pm.003}$ &
    $0.822^{\pm.003}$ &
    $0.075^{\pm.002}$ &
    $2.826^{\pm.011}$ & $\mathbf{9.517}^{\pm.048}$ & $2.076^{\pm.052}$ \\

    \midrule

    \checkmark & \checkmark & \checkmark &
    $\mathbf{0.550}^{\pm.003}$ &
    $\mathbf{0.744}^{\pm.003}$ &
    $\mathbf{0.836}^{\pm.003}$ &
    $\mathbf{0.035}^{\pm.002}$ &
    $\mathbf{2.779}^{\pm.006}$ & ${9.534}^{\pm.066}$ & $\mathbf{2.155}^{\pm.046}$ \\

    \bottomrule
  \end{tabular}
  \label{tab:tables10}
\end{table*}

\begin{table}[t]
\caption{Comparison of computational complexity. Part-level evaluation reports the average performance of arms and legs.}
\centering
\resizebox{\columnwidth}{!}{
\begin{tabular}{lccccc}
\toprule
\multirow{2}{*}{Method} & \multirow{2}{*}{\#Params} & \multirow{2}{*}{AIT} 
& \multicolumn{3}{c}{Part-level Evaluation (Avg.)} \\
\cmidrule(lr){4-6}
& & & R-Precision (Top-3)$\uparrow$ & FID$\downarrow$ & MM-Dist$\downarrow$ \\
\midrule
T2M-GPT~\cite{zhang2023generating} & 247.5M & 277ms & $0.706^{\pm.003}$ & $0.198^{\pm.003}$ & $3.534^{\pm.008}$  \\
ParCo~\cite{zou2024parco} & 168.4M & 643ms & $0.733^{\pm.003}$ & $0.166^{\pm.003}$ & $3.404^{\pm.010}$  \\
MoMask~\cite{guo2024momask} & 44.8M & 80ms & $0.724^{\pm.003}$ & $0.139^{\pm.003}$ & $3.476^{\pm.007}$ \\
Ours & 78.3M & 158ms & $0.778^{\pm.003}$ & $0.105^{\pm.003}$ & $3.100^{\pm.007}$  \\
\bottomrule
\end{tabular}
}
\label{tab:tables11}
\end{table}

\section{Additional Quantitative Results}\label{sec:E}
All experiments were performed on the HumanML3D~\cite{guo2022generating} dataset, except for Tab.~\ref{tab:tables9}.

\subsection{Detailed Results \& Ablation Studies}\label{sec:E.1}
We provide quantitative comparisons including additional models in Tab. ~\ref{tab:tables9}. Detailed ablation study results for the proposed components are presented in Tab.~\ref{tab:tables10}. Note that PTG is fed into the part transformer and is thus related to its output; when used alone without PG and HPF, this output cannot be reflected. Therefore, this case was excluded from the ablation study.

\subsection{Computational Complexity}\label{sec:E.2}
We report computational complexity, including the number of parameters and average inference time (AIT), in Tab.~\ref{tab:tables11}. Notably, ParTY reduces the parameter count by more than 2× and improves AIT by over 4× compared to ParCo~\cite{zou2024parco}, a conventional part-wise method, while also enhancing part expressiveness. Although ParTY has slightly higher complexity than MoMask due to the use of part transformers, it achieves substantially higher part expressiveness, representing a reasonable trade-off.

\subsection{Temporal-aware VQ-VAE}\label{sec:E.3}
\subsubsection{Porting to Additional Model}\label{sec:E.3.1}
To demonstrate the generality of our Temporal-aware VQ-VAE, we report the results of porting it to an additional model, ParCo~\cite{zou2024parco}, in Tab.~\ref{tab:tables12}.

\begin{table}[t]
\caption{Porting Temporal-aware VQ-VAE to ParCo~\cite{zou2024parco}. AIT is averaged over 100 samples on an RTX A5000 GPU.}
\centering
\resizebox{\columnwidth}{!}{
\renewcommand{\arraystretch}{1.3}
\setlength{\tabcolsep}{8pt}
\begin{tabular}{lccccccc}
\toprule
\multirow{2}{*}{Method} & \multirow{2}{*}{\makecell{Window\\size}} & \multicolumn{3}{c}{Reconstruction} & \multicolumn{2}{c}{Generation} \\
\cmidrule(lr){3-5} \cmidrule(lr){6-7}
& & \# Params & FID$\downarrow$ & MPJPE$\downarrow$ & FID$\downarrow$ & AIT \\
\midrule
ParCo & 4 & 6.35M & 0.021 & 0.108 & 0.109 & \multirow{2}{*}{65ms} \\
\cellcolor{blue!15}\hspace{0.5em}+ Ours & \cellcolor{blue!15}4 & \cellcolor{blue!15}7.93M & \cellcolor{blue!15}0.005 {\scriptsize \textcolor{blue}{(+76\%)}} & \cellcolor{blue!15}0.034 {\scriptsize \textcolor{blue}{(+68\%)}} & \cellcolor{blue!15}0.077 {\scriptsize \textcolor{blue}{(+29\%)}} & \\
\midrule
ParCo & 8 & 4.42M & 0.047 & 0.166 & 0.172 & \multirow{2}{*}{\makecell{33ms\\{\scriptsize \textcolor{red}{(+49\%)}}}} \\
\cellcolor{blue!15}\hspace{0.5em}+ Ours & \cellcolor{blue!15}8 & \cellcolor{blue!15}6.01M & \cellcolor{blue!15}0.009 {\scriptsize \textcolor{blue}{(+81\%)}} & \cellcolor{blue!15}0.041 {\scriptsize \textcolor{blue}{(+75\%)}} & \cellcolor{blue!15}0.085 {\scriptsize \textcolor{blue}{(+50\%)}} & \\
\midrule
ParCo & 12 & 3.59M & 0.090 & 0.243 & 0.255 & \multirow{2}{*}{\makecell{23ms\\{\scriptsize \textcolor{red}{(+64\%)}}}} \\
\cellcolor{blue!15}\hspace{0.5em}+ Ours & \cellcolor{blue!15}12 & \cellcolor{blue!15}5.18M & \cellcolor{blue!15}0.017 {\scriptsize \textcolor{blue}{(+80\%)}} & \cellcolor{blue!15}0.060 {\scriptsize \textcolor{blue}{(+75\%)}} & \cellcolor{blue!15}0.098 {\scriptsize \textcolor{blue}{(+61\%)}} & \\
\bottomrule
\end{tabular}
}
\label{tab:tables12}
\end{table}

\subsubsection{Ablation Studies}\label{sec:E.3.2}
We conducted ablation studies on the proposed components (LTE, GTE) in Temporal-aware VQ-VAE and report the reconstruction performance in Tab.~\ref{tab:tables13}.

\begin{table}[t]
\caption{Ablation studies on proposed components in Temporal-aware VQ-VAE.}
\centering
\small
\begin{tabular}{@{}ccccc@{}}
\toprule
LTE & GTE & FID $\downarrow$ & MPJPE $\downarrow$ \\
\midrule
& & $0.038^{\pm.000}$ & $0.121^{\pm.000}$ \\
\checkmark & & $0.018^{\pm.000}$ & $0.049^{\pm.000}$ \\
& \checkmark & $0.020^{\pm.000}$ & $0.071^{\pm.000}$ \\
\checkmark & \checkmark & $0.007^{\pm.000}$ & $0.019^{\pm.000}$ \\
\bottomrule
\end{tabular}
\label{tab:tables13}
\end{table}

\begin{table}[t]
\caption{Quantitative results according to the number of embeddings in PTG.}
\centering
\resizebox{\columnwidth}{!}{
\large
\begin{tabular}{cccccc}
\toprule
\multirow{2}{*}{\# of Emb.} & \multicolumn{3}{c}{R-Precision $\uparrow$} & \multirow{2}{*}{FID $\downarrow$} & \multirow{2}{*}{MM-Dist $\downarrow$} \\
\cmidrule(lr){2-4}
& Top-1 & Top-2 & Top-3 & & \\
\midrule
1 & $0.531^{\pm.003}$ & $0.722^{\pm.003}$ & $0.815^{\pm.003}$ & $0.060^{\pm.003}$ & $2.842^{\pm.011}$ \\  
2 & $0.538^{\pm.003}$ & $0.731^{\pm.002}$ & $0.827^{\pm.003}$ & $0.052^{\pm.003}$ & $2.825^{\pm.015}$ \\
3 & $0.541^{\pm.002}$ & $0.735^{\pm.003}$ & $0.829^{\pm.003}$ & $0.047^{\pm.003}$ & $2.811^{\pm.007}$ \\
4 & $\mathbf{0.550}^{\pm.003}$ & $\mathbf{0.744}^{\pm.003}$ & $\mathbf{0.836}^{\pm.003}$ & $\underline{0.035}^{\pm.002}$ & $\mathbf{2.779}^{\pm.006}$ \\
5 & $\underline{0.546}^{\pm.003}$ & $\underline{0.739}^{\pm.002}$ & $\underline{0.833}^{\pm.003}$ & $\mathbf{0.033}^{\pm.002}$ & $\underline{2.804}^{\pm.011}$ \\
6 & $0.531^{\pm.003}$ & $0.727^{\pm.003}$ & $0.819^{\pm.003}$ & $0.045^{\pm.003}$ & $2.878^{\pm.014}$ \\
\bottomrule
\end{tabular}
}
\label{tab:tables14}
\end{table}

\subsection{Part-aware Text Grounding}\label{sec:E.4}
\subsubsection{Analysis of the Number of Embeddings}\label{sec:E.4.1}
We conducted experiments under various conditions to find the optimal number of embeddings $K$ in Part-aware Text Grounding (PTG). As shown in Tab.~\ref{tab:tables14}, we tested cases with 1, 2, 3, 4, 5, and 6 embeddings. The results showed that 4 embeddings achieved the best R-Precision performance, while 5 views showed slightly lower R-Precision but the highest FID performance. From 1 to 4 embeddings, we observed a general improvement in performance, indicating that adequate representation through a sufficient number of embeddings is necessary for properly aligning text descriptions to each part. However, the performance decline with 6 embeddings suggests that simply increasing the number of embeddings is not beneficial. This can be attributed to potential role overlap between embeddings and increased complexity in the Part Gate's weighting process when the number of embeddings exceeds a certain threshold.

\subsubsection{Ablation Studies on Auxilary Loss}\label{sec:E.4.2}
Tab.~\ref{tab:tables15} reports ablation study results on the auxiliary loss, and Tab.~\ref{tab:tables16} presents the corresponding results for part-level evaluation. These results demonstrate that using LLM-generated part text as auxiliary supervision provides modest improvements to text-part alignment.

\begin{table}[t]
\caption{Ablation studies on auxilary loss.}
\centering
\resizebox{\columnwidth}{!}{
\large
\begin{tabular}{cccccc}
\toprule
\multirow{2}{*}{Method} & \multicolumn{3}{c}{R-Precision $\uparrow$} & \multirow{2}{*}{FID $\downarrow$} & \multirow{2}{*}{MM-Dist $\downarrow$} \\
\cmidrule(lr){2-4}
& Top-1 & Top-2 & Top-3 & & \\
\midrule
Ours (Full model) & $0.550^{\pm.003}$ & $0.744^{\pm.003}$ & $0.836^{\pm.003}$ & $0.035^{\pm.002}$ & $2.779^{\pm.006}$ \\  
w/o aux. loss & $0.543^{\pm.003}$ & $0.739^{\pm.003}$ & $0.830^{\pm.003}$ & $0.042^{\pm.003}$ & $2.805^{\pm.011}$ \\
\bottomrule
\end{tabular}
}
\label{tab:tables15}
\end{table}

\begin{table}[t]
\caption{Ablation studies on auxilary loss with \textbf{part-level evaluation} metrics.}
\centering
\resizebox{\columnwidth}{!}{
\renewcommand{\arraystretch}{1.3}
\begin{tabular}{@{}llcccc@{}}
\toprule
Method & Part & \makecell{R-Precision\\(Top-1)$\uparrow$} & \makecell{R-Precision\\(Top-3)$\uparrow$} & FID$\downarrow$ & MM-Dist$\downarrow$ \\
\midrule
\multirow{2}{*}{Ours (Full model)} & Arms & $0.506^{\pm.003}$ & $0.802^{\pm.002}$ & $0.133^{\pm.002}$ & $3.079^{\pm.005}$ \\
& Legs & $0.463^{\pm.003}$ & $0.755^{\pm.003}$ & $0.078^{\pm.003}$ & $3.122^{\pm.008}$ \\
\midrule
\multirow{2}{*}{w/o aux. loss} & Arms & $0.491^{\pm.003}$ & $0.793^{\pm.002}$ & $0.148^{\pm.003}$ & $3.098^{\pm.008}$ \\
& Legs & $0.444^{\pm.002}$ & $0.732^{\pm.003}$ & $0.092^{\pm.003}$ & $3.167^{\pm.008}$ \\
\bottomrule
\end{tabular}
}
\label{tab:tables16}
\end{table}

\subsubsection{LLM Prompt Details}\label{sec:E.4.3}
\begin{figure}
    \centering
    \includegraphics[width=1\linewidth]{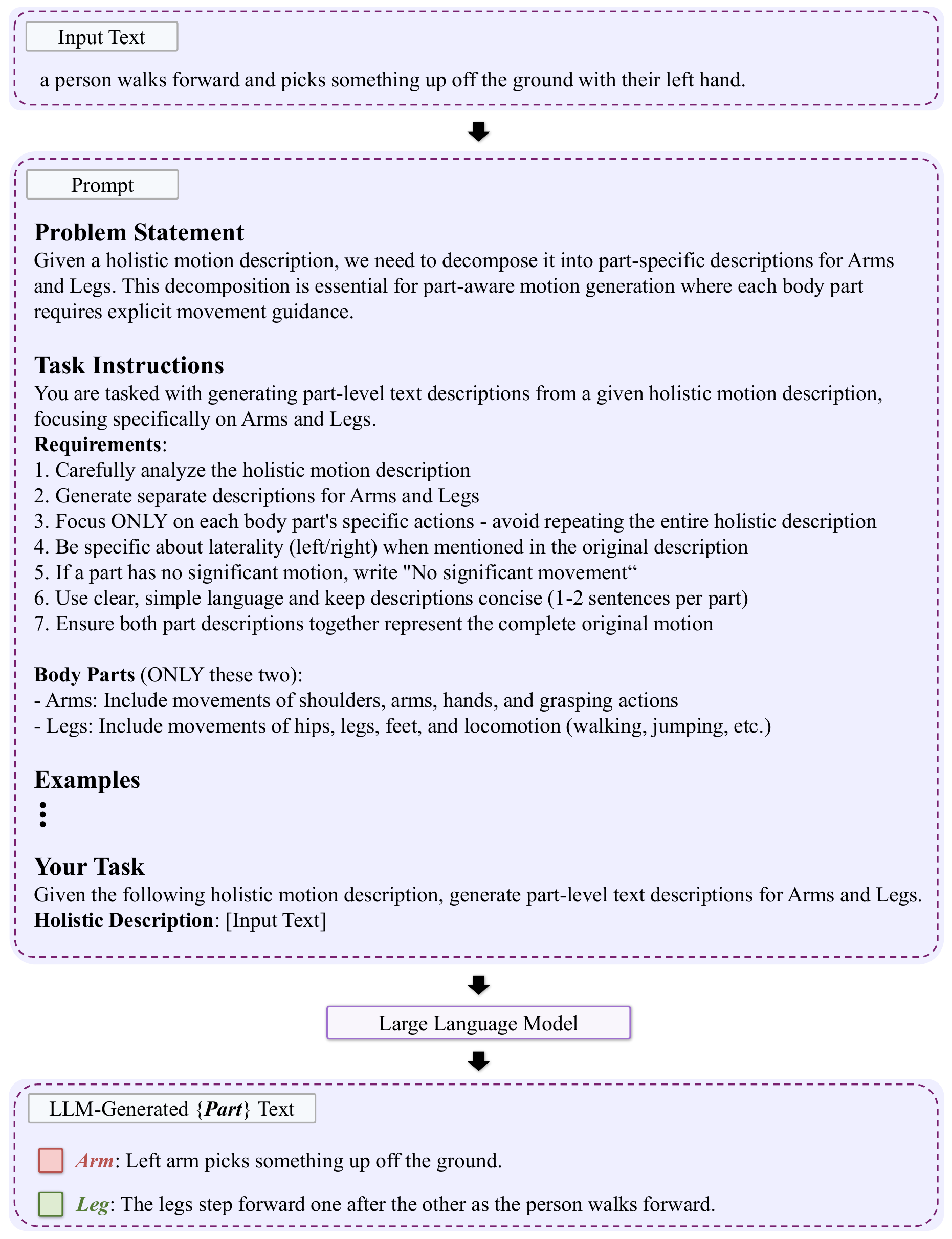}
    \caption{LLM prompt details.}
    \label{fig:sup_llm}
\end{figure}

We provide the complete prompt used to generate part text descriptions with an LLM in Fig.~\ref{fig:sup_llm}. We employ \texttt{gemini-2.5-flash} as our LLM, which takes a holistic motion description as input and generates corresponding part-level descriptions for arms and legs.  When a part description indicates ``No significant movement,'' we use the holistic text embedding as auxiliary supervision for that part instead of the LLM-generated part text. The \textbf{Examples} in Fig.~\ref{fig:sup_llm} were constructed as follows:

\begin{itemize}
    \item \textbf{Example 1:} \\
    \textit{Holistic Description:} ``A person waves with their right hand while standing still.'' \\
    \textit{Part-level Descriptions:}
    \begin{itemize}
        \item Arms: Right arm raises up and waves side to side.
        \item Legs: No significant movement.
    \end{itemize}
    
    \item \textbf{Example 2:} \\
    \textit{Holistic Description:} ``A person runs forward quickly.'' \\
    \textit{Part-level Descriptions:}
    \begin{itemize}
        \item Arms: Both arms swing back and forth rhythmically to support the running motion.
        \item Legs: The legs alternate rapidly, pushing off the ground to propel the body forward in a running motion.
    \end{itemize}
    
    \item \textbf{Example 3:} \\
    \textit{Holistic Description:} ``A person jumps up and raises both arms above their head.'' \\
    \textit{Part-level Descriptions:}
    \begin{itemize}
        \item Arms: Both arms lift upward and extend above the head.
        \item Legs: The legs push off the ground forcefully to jump, then land back down.
    \end{itemize}
    
    \item \textbf{Example 4:} \\
    \textit{Holistic Description:} ``A person sits down on a chair.'' \\
    \textit{Part-level Descriptions:}
    \begin{itemize}
        \item Arms: No significant movement.
        \item Legs: The legs bend at the knees and lower the body into a sitting position.
    \end{itemize}
    
    \item \textbf{Example 5:} \\
    \textit{Holistic Description:} ``A person kicks a ball with their right foot.'' \\
    \textit{Part-level Descriptions:}
    \begin{itemize}
        \item Arms: No significant movement.
        \item Legs: The right leg swings forward to kick, while the left leg supports the body's weight.
    \end{itemize}
\end{itemize}

\subsubsection{Quality Evaluation of Generated Text}\label{sec:E.4.4}
We generated part text descriptions multiple times using the same prompt and measured performance using them as auxiliary supervision, with results reported in Tab.~\ref{tab:tables17}. Although the LLM-generated text varies slightly across generations, the performance remains stable with minimal fluctuation due to the small weight of the auxiliary loss.

\begin{table}[t]
\caption{Quantitative results with multiple LLM-generated part texts.}
\centering
\resizebox{\columnwidth}{!}{
\large
\begin{tabular}{cccccc}
\toprule
\multirow{2}{*}{Types} & \multicolumn{3}{c}{R-Precision $\uparrow$} & \multirow{2}{*}{FID $\downarrow$} & \multirow{2}{*}{MM-Dist $\downarrow$} \\
\cmidrule(lr){2-4}
& Top-1 & Top-2 & Top-3 & & \\
\midrule
$\text{LLM}_1$ & $0.549^{\pm.003}$ & $0.744^{\pm.002}$ & $\mathbf{0.836}^{\pm.003}$ & $0.035^{\pm.002}$ & $\mathbf{2.777}^{\pm.009}$ \\  
$\text{LLM}_2$ & $\mathbf{0.551}^{\pm.003}$ & $0.743^{\pm.003}$ & $0.835^{\pm.003}$ & $0.038^{\pm.003}$ & $2.792^{\pm.009}$ \\  
$\text{LLM}_3$ & $0.550^{\pm.003}$ & $0.744^{\pm.003}$ & $\mathbf{0.836}^{\pm.002}$ & $0.035^{\pm.002}$ & $2.779^{\pm.006}$ \\  
$\text{LLM}_4$ & $0.550^{\pm.003}$ & $\mathbf{0.745}^{\pm.003}$ & $\mathbf{0.836}^{\pm.003}$ & $0.036^{\pm.003}$ & $2.788^{\pm.007}$ \\  
$\text{LLM}_5$ & $0.548^{\pm.003}$ & $0.743^{\pm.003}$ & $\mathbf{0.836}^{\pm.002}$ & $\mathbf{0.032}^{\pm.003}$ & $2.790^{\pm.011}$ \\  
\bottomrule
\end{tabular}
}
\label{tab:tables17}
\end{table}

\subsubsection{Group of Text Descriptions}\label{sec:E.4.5}
We provide the group of text descriptions mentioned in manuscript Fig. 6 related to Part-aware Text Grounding. This group consists of 30 \textit{``squat"}-related text descriptions, constructed as follows: we extracted all text descriptions containing \textit{``squat"} from the entire HumanML3D dataset, encoded them into features using the CLIP~\cite{radford2021learning} text encoder, and selected the text description \textit{``a person squats down"} along with 29 text descriptions whose features are closest to this anchor description, resulting in a total of 30 descriptions. The following is the list of text descriptions:

\begin{enumerate}
    \item \textit{``a person squats down."}
    \item \textit{``a man is doing squats."}
    \item \textit{``a person is squatting down."}
    \item \textit{``a person is squatting down."}
    \item \textit{``the person is doing squats."}
    \item \textit{``someone is squatting down."}
    \item \textit{``a person squatting, raises arms."}
    \item \textit{``a person slightly squats down."}
    \item \textit{``a person lifting weights or squatting."}
    \item \textit{``a person performs a single squat."}
    \item \textit{``a person does an exercise squat."}
    \item \textit{``a person raises his hands, squats."}
    \item \textit{``a person in squat position while extending elbows."}
    \item \textit{``a person is squatting while moving their hands."}
    \item \textit{``a person squats down and then stands back up."}
    \item \textit{``a person squats down and holds out their arms."}
    \item \textit{``a squatting person raises their arms upwards from their sides."}
    \item \textit{``a person holds something above their shoulders and squats slightly."}
    \item \textit{``a person squats down and puts their hands above their head."}
    \item \textit{``a person does a squat and raises both arms over its head."}
    \item \textit{``a figure squatting whilst hold both arms out towards the front."}
    \item \textit{``the person did a squat and they raised both arms forward."}
    \item \textit{``person squats down repeatedly with arms raised up to hold weights."}
    \item \textit{``a person squats down and moves their hands around at head level."}
    \item \textit{``person is standing in a slight squat position with hands resting on thighs."}
    \item \textit{``person is squatting and raises both arms up straight out and then down."}
    \item \textit{``a person squats down and puts their hands up to their face while squatting."}
    \item \textit{``the person puts their hand on their face then squats down like they're going underwater."}
    \item \textit{``a person raised their arms above their head, and squatted."}
    \item \textit{``a person squats by bending both knees and both elbows and moving arms above legs without touching them."}
\end{enumerate}

\subsection{Part Guidance}\label{sec:E.5}
\subsubsection{Analysis of the Size of Window}\label{sec:E.5.1}
We evaluate quantitative performance using Part Guidance generated with different window sizes and report the results in Tab.~\ref{tab:tables18}. Performance increases progressively from window size 1 to 3, then slightly decreases from window size 4 onward. This suggests that a window size of 3 provides the optimal amount of future part information for the model to utilize effectively, while window sizes of 4 or larger introduce additional complexity that negatively impacts performance.

\begin{table}[t]
\caption{Quantitative results for different window sizes in PG.}
\centering
\resizebox{\columnwidth}{!}{
\large
\begin{tabular}{cccccc}
\toprule
\multirow{2}{*}{\makecell{Window\\size}} & \multicolumn{3}{c}{R-Precision $\uparrow$} & \multirow{2}{*}{FID $\downarrow$} & \multirow{2}{*}{MM-Dist $\downarrow$} \\
\cmidrule(lr){2-4}
& Top-1 & Top-2 & Top-3 & & \\
\midrule
1 & $0.545^{\pm.003}$ & $0.735^{\pm.003}$ & $0.829^{\pm.003}$ & $0.051^{\pm.003}$ & $2.822^{\pm.012}$ \\  
2 & $0.547^{\pm.003}$ & $0.740^{\pm.003}$ & $0.833^{\pm.002}$ & $0.040^{\pm.003}$ & $2.801^{\pm.008}$ \\  
3 & $0.550^{\pm.003}$ & $\mathbf{0.744}^{\pm.003}$ & $\mathbf{0.836}^{\pm.002}$ & $\mathbf{0.035}^{\pm.002}$ & $2.779^{\pm.006}$ \\   
4 & $\mathbf{0.551}^{\pm.003}$ & $0.741^{\pm.003}$ & $0.835^{\pm.003}$ & $0.036^{\pm.002}$ & $\mathbf{2.754}^{\pm.009}$ \\
5 & $0.548^{\pm.003}$ & $0.740^{\pm.002}$ & $0.833^{\pm.003}$ & $0.039^{\pm.003}$ & $2.793^{\pm.007}$ \\  
\bottomrule
\end{tabular}
}
\label{tab:tables18}
\end{table}

\begin{figure}[t]
    \centering
    \includegraphics[width=1\linewidth]{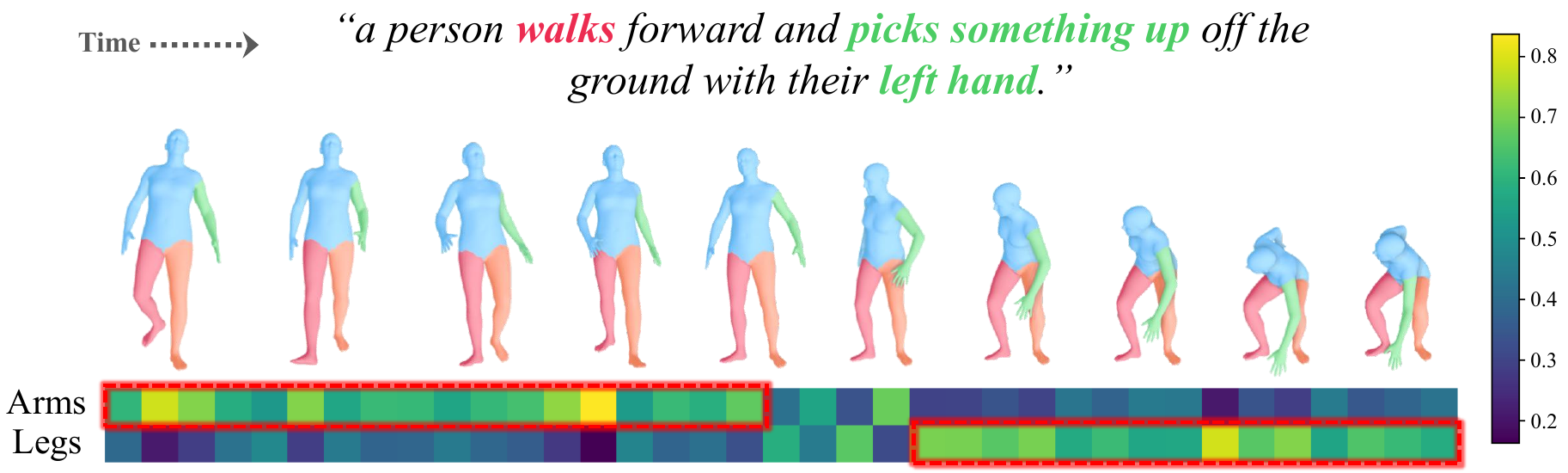}
    \caption{Visualization of cross attention map of HPF.}
    \label{fig:sup_hpf}
\end{figure}

\begin{figure}[t]
    \centering
    \includegraphics[width=1\linewidth]{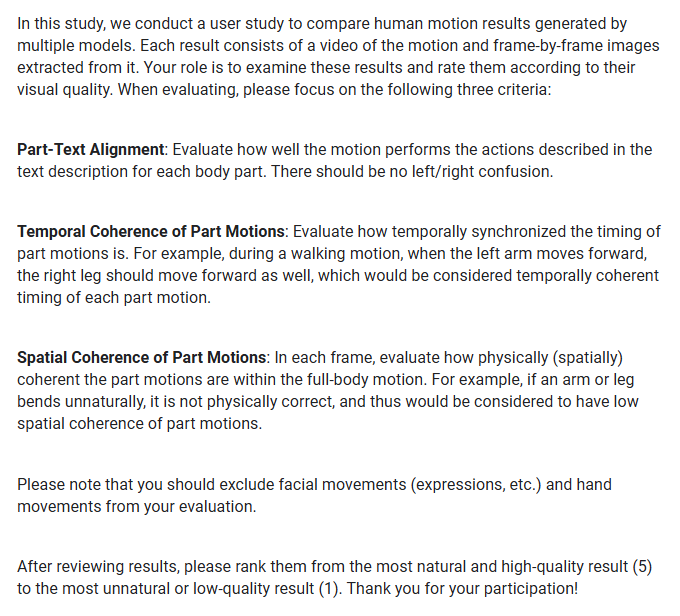}
    \vspace{-6mm}
    \caption{Guidelines for user study.}
    \label{fig:sup_user_1}
\end{figure}

\begin{figure}[t]
    \centering
    \includegraphics[width=1\linewidth]{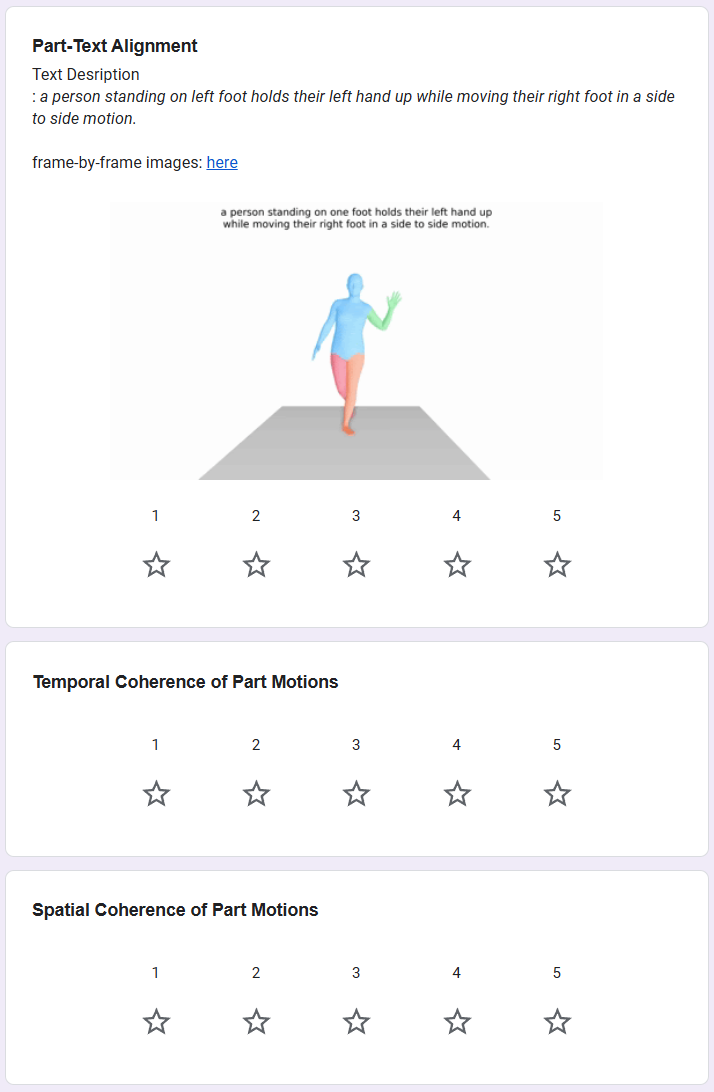}
    \caption{User evaluation interface for the user study.}
    \vspace{-5mm}
    \label{fig:sup_user_2}
\end{figure}

\begin{figure*}[t]
\centering
\begin{subfigure}[b]{0.32\textwidth}
    \centering
    \includegraphics[width=\textwidth]{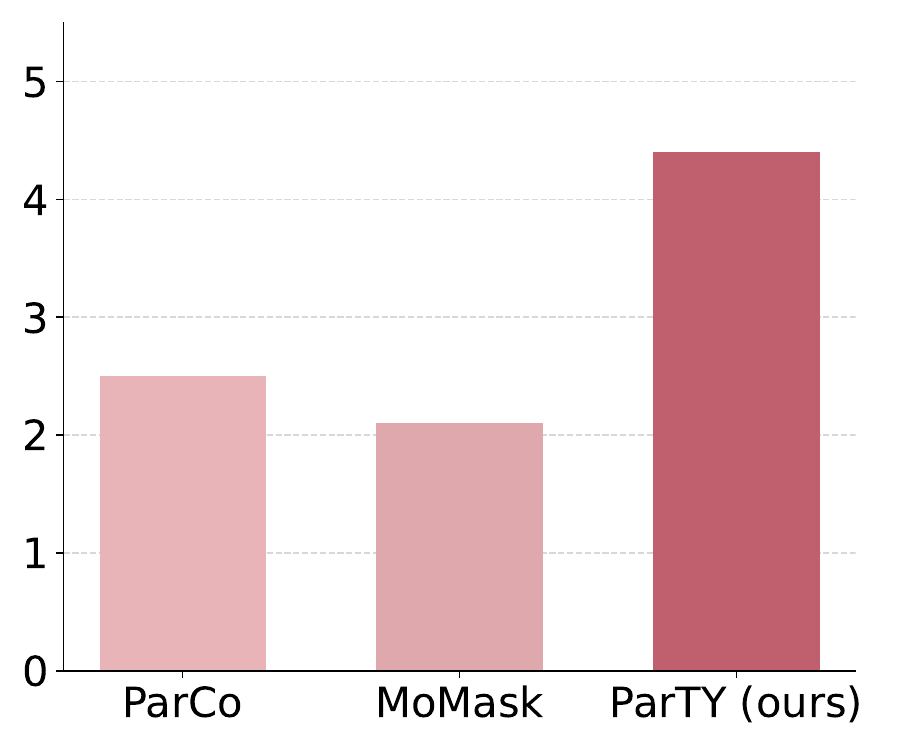}
    \caption{Part-Text Alignment}
\end{subfigure}
\hfill
\begin{subfigure}[b]{0.32\textwidth}
    \centering
    \includegraphics[width=\textwidth]{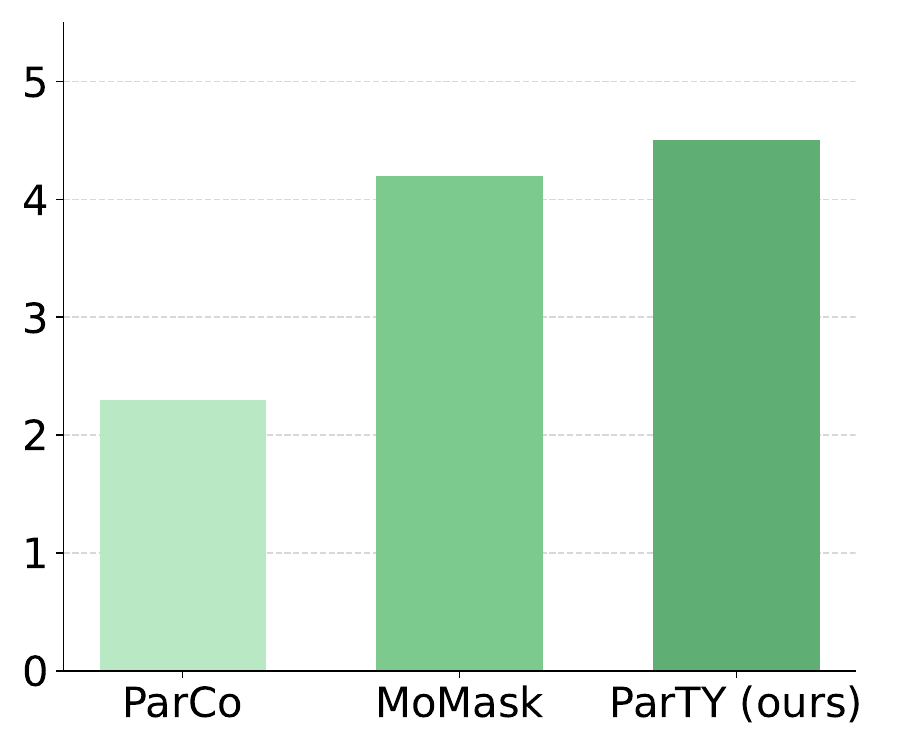}
    \caption{Temporal Coherence of Part Motions}
\end{subfigure}
\hfill
\begin{subfigure}[b]{0.32\textwidth}
    \centering
    \includegraphics[width=\textwidth]{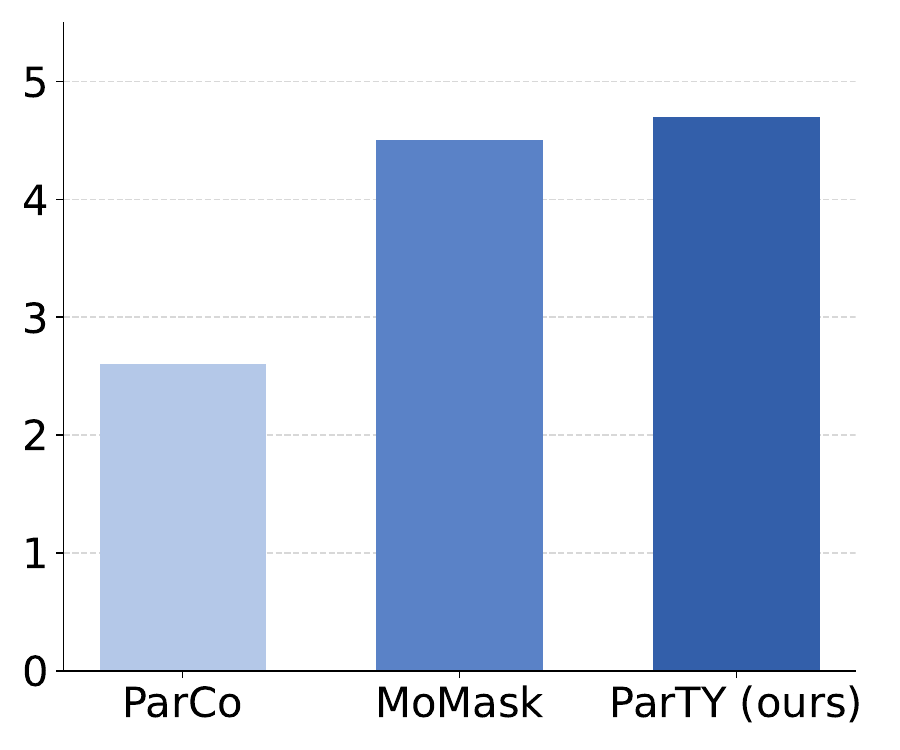}
    \caption{Spatial Coherence of Part Motions}
\end{subfigure}
\caption{User study results on HumanML3D dataset. Each bar represents the average score on a scale from 1 to 5.}
\label{fig:sup_user_result}
\end{figure*}

\subsection{Holistic-Part Fusion}\label{sec:E.6}

\subsubsection{Additional Visualization of Attention Map}\label{sec:E.6.1}
We provide additional sample of cross-attention map visualization from the Holistic-Part Fusion process in Fig.~\ref{fig:sup_hpf}.

\section{User Study Details}\label{sec:F}
We conducted a user study to validate our proposed model and metrics. As described in Fig.~\ref{fig:sup_user_1}, we provided participants with guidelines for evaluating (1) Part-Text Alignment, (2) Temporal Coherence of Part Motion, and (3) Spatial Coherence of Part Motion. Participants assessed motions generated by MoMask~\cite{guo2024momask}, ParCo~\cite{zou2024parco}, and ParTY (ours), assigning scores for each metric as shown in Fig.~\ref{fig:sup_user_2}.

Fig.~\ref{fig:sup_user_result} presents the results of our user study with 50 participants. The results in (a) confirm that ParTY achieves superior part expressiveness compared to other methods, consistent with human visual perception. Furthermore, (b) and (c) demonstrate that ParTY maintains stable coherence and that our proposed coherence-level metrics effectively capture aspects of motion quality that align with human judgment.

% \section{Limitations}\label{sec:G}
% While our method demonstrates substantial improvements over existing approaches, certain limitations remain. A key constraint lies in the limited window size of Part Guidance. Presently, we construct Part Guidance by aggregating tokens generated by the part transformer across 3 time steps. While our results validate the utility of incorporating future information, there exists potential for further improvement through larger window sizes that capture information from more distant future frames. However, the current Part Guidance employs simple token addition without explicit temporal modeling, resulting in loss of temporal structure. Addressing this limitation by incorporating temporal-aware aggregation in Part Guidance construction could enable the effective use of larger window sizes, potentially yielding more stable and consistent motion generation.

\newpage

\begin{table*}[t]
\caption{Quantitative comparison on HumanML3D and KIT-ML. \textbf{Bold} indicates the best result, while \underline{underlined} refers the second-best. The right arrow $\rightarrow$ indicates that closer values to ground truth are preferred.}
\small
\resizebox{\linewidth}{!}{%
\begin{tabular}{@{}clccccccc@{}}
\toprule
\multirow{2}{*}{Datasets} & \multirow{2}{*}{Method} & \multicolumn{3}{c}{R Precision $\uparrow$}                                                                                                                & \multicolumn{1}{c}{\multirow{2}{*}{FID$\downarrow$}} & \multirow{2}{*}{MM-Dist$\downarrow$}              & \multirow{2}{*}{Diversity$\rightarrow$}           & \multirow{2}{*}{MultiModality$\uparrow$}              \\ \cmidrule(lr){3-5}
             ~& ~& \multicolumn{1}{c}{Top 1} & \multicolumn{1}{c}{Top 2} & \multicolumn{1}{c}{Top 3} & \multicolumn{1}{c}{}                     &                          &                            &                            \\ \midrule
\multirow{21}{*}{\makecell[c]{HumanML3D}}  &  Real motion & 
 $0.511^{\pm.003}$ &
 $0.703^{\pm.003}$ &
 $0.797^{\pm.002}$ &
 $0.002^{\pm.000}$ &
 $2.974^{\pm.008}$ &
 $9.503^{\pm.065}$ &
  - \\
 \cline{2-9} \vspace{-0.3cm} \\ 

~& TM2T \cite{guo2022tm2t}&
  $0.424^{\pm.003}$ &
  $0.618^{\pm.003}$ &
  $0.729^{\pm.002}$ &
  $1.501^{\pm.017}$ &
  $3.467^{\pm.011}$ &
  $8.589^{\pm.076}$ &
  $2.424^{\pm.093}$ \\
  
~& T2M \cite{guo2022generating}&
  $0.457^{\pm.002}$ &
  $0.639^{\pm.003}$ &
  $0.740^{\pm.003}$ &
  $1.067^{\pm.002}$ &
  $3.340^{\pm.008}$ &
  $9.188^{\pm.002}$ &
  $2.090^{\pm.083}$ \\

~& MDM \cite{tevet2022human}&
  $0.320^{\pm.005}$ &
  $0.498^{\pm.004}$ &
  $0.611^{\pm.007}$ &
  ${0.544}^{\pm.044}$ &
  $5.566^{\pm.027}$ &
  ${9.559}^{\pm.086}$ &
  $\mathbf{2.799}^{\pm.072}$ \\

~& MLD \cite{chen2023executing} &
  ${0.481}^{\pm.003}$ &
  ${0.673}^{\pm.003}$ &
  ${0.772}^{\pm.002}$ &
  ${0.473}^{\pm.013}$ &
  ${3.196}^{\pm.010}$ &
  $9.724^{\pm.082}$ &
  ${2.413}^{\pm.079}$ \\

~& T2M-GPT \cite{zhang2023generating} &
    ${0.491}^{\pm.003}$ &
    ${0.680}^{\pm.003}$ &
    ${0.775}^{\pm.002}$ &
    ${0.116}^{\pm.004}$ &
    ${3.118}^{\pm.011}$ &
    ${9.761}^{\pm.081}$ &
    $1.856^{\pm.011}$ \\ 

~& AttT2M \cite{zhong2023attt2m} &
    ${0.499}^{\pm.003}$ &
    ${0.690}^{\pm.002}$ &
    ${0.786}^{\pm.002}$ &
    ${0.112}^{\pm.006}$ &
    ${3.038}^{\pm.007}$ &
    ${9.700}^{\pm.090}$ &
    ${2.452}^{\pm.051}$ \\ 

~& ParCo \cite{zou2024parco} &
    ${0.515}^{\pm.003}$ &
    ${0.706}^{\pm.003}$ &
    ${0.801}^{\pm.002}$ &
    ${0.109}^{\pm.005}$ &
    ${2.927}^{\pm.008}$ &
    ${9.576}^{\pm.088}$ &
    $1.382^{\pm.060}$ \\

~& ReMoDiffuse \cite{zhang2023remodiffuse} &
    ${0.510}^{\pm.005}$ &
    ${0.698}^{\pm.006}$ &
    ${0.795}^{\pm.004}$ &
    ${0.103}^{\pm.004}$ &
    ${2.974}^{\pm.016}$ &
    ${9.018}^{\pm.075}$ &
    $1.795^{\pm.043}$ \\ 

~& MMM \cite{pinyoanuntapong2024mmm} &
    ${0.504}^{\pm.003}$ &
    ${0.696}^{\pm.003}$ &
    ${0.794}^{\pm.002}$ &
    ${0.080}^{\pm.003}$ &
    ${2.998}^{\pm.007}$ &
    ${9.411}^{\pm.058}$ &
    $1.164^{\pm.041}$ \\ 

 ~& SALAD \cite{hong2025salad} &
    ${\textbf{0.581}}^{\pm.003}$ &
    ${\textbf{0.769}}^{\pm.003}$ &
    ${\textbf{0.857}}^{\pm.002}$ &
    ${0.076}^{\pm.002}$ &
    ${\textbf{2.649}}^{\pm.009}$ &
    ${9.696}^{\pm.096}$ &
    $1.751^{\pm.062}$ \\

~& BAD \cite{hosseyni2025bad} &
    ${0.517}^{\pm.002}$ &
    ${0.713}^{\pm.003}$ &
    ${0.808}^{\pm.003}$ &
    ${0.065}^{\pm.003}$ &
    ${2.901}^{\pm.008}$ &
    ${9.694}^{\pm.068}$ &
    $1.194^{\pm.044}$ \\

 ~& BAMM \cite{pinyoanuntapong2024bamm} &
    ${0.525}^{\pm.002}$ &
    ${0.720}^{\pm.003}$ &
    ${0.814}^{\pm.003}$ &
    ${0.055}^{\pm.002}$ &
    ${2.919}^{\pm.008}$ &
    ${9.717}^{\pm.089}$ &
    $1.687^{\pm.051}$ \\

 ~& MoMask \cite{guo2024momask} &
    ${0.521}^{\pm.002}$ &
    ${0.713}^{\pm.002}$ &
    ${0.807}^{\pm.002}$ &
    ${0.045}^{\pm.002}$ &
    ${2.958}^{\pm.008}$ &
    - &
    $1.241^{\pm.040}$ \\

 ~& Light-T2M \cite{zeng2025light} &
    ${0.511}^{\pm.003}$ &
    ${0.699}^{\pm.002}$ &
    ${0.795}^{\pm.002}$ &
    ${0.040}^{\pm.002}$ &
    ${3.002}^{\pm.008}$ &
    - &
    $1.670^{\pm.061}$ \\

 ~& MoGenTS \cite{yuan2024mogents} &
    ${0.529}^{\pm.003}$ &
    ${0.719}^{\pm.002}$ &
    ${0.812}^{\pm.002}$ &
    ${0.033}^{\pm.001}$ &
    ${2.867}^{\pm.006}$ &
    ${9.570}^{\pm.077}$ &
    - \\

 ~& LAMP \cite{li2024lamp} &
    ${\underline{0.557}}^{\pm.003}$ &
    ${\underline{0.751}}^{\pm.002}$ &
    ${\underline{0.843}}^{\pm.001}$ &
    ${0.032}^{\pm.002}$ &
    ${\underline{2.759}}^{\pm.007}$ &
    ${9.571}^{\pm.069}$ &
    - \\

 ~& DisCoRD \cite{cho2025discord} &
    ${0.524}^{\pm.003}$ &
    ${0.715}^{\pm.003}$ &
    ${0.809}^{\pm.002}$ &
    ${0.032}^{\pm.002}$ &
    ${2.938}^{\pm.010}$ &
    - &
    $1.288^{\pm.043}$ \\

 ~& BiPO \cite{hong2026bipo} & 
    ${0.523}^{\pm.003}$ &
    ${0.714}^{\pm.002}$ &
    ${0.809}^{\pm.002}$ &
    ${\underline{0.030}}^{\pm.002}$ &
    ${2.880}^{\pm.009}$ &
    ${9.556}^{\pm.076}$ &
    $1.374^{\pm.047}$ \\

 ~& Motion Anything \cite{zhang2025motion} &
    ${0.546}^{\pm.003}$ &
    ${0.735}^{\pm.002}$ &
    ${0.829}^{\pm.002}$ &
    $\textbf{0.028}^{\pm.005}$ &
    ${2.859}^{\pm.010}$ &
    ${\textbf{9.521}}^{\pm.083}$ &
    $\underline{2.705}^{\pm.068}$ \\    

\cline{2-9} \vspace{-0.3cm} \\ 
~& \textbf{ParTY (Ours)} &
    ${0.550}^{\pm.003}$ &
    ${0.744}^{\pm.003}$ &
    ${0.836}^{\pm.003}$ &
    ${0.035}^{\pm.002}$ &
    ${2.779}^{\pm.006}$ &
    $\underline{9.534}^{\pm.066}$ &
    ${2.155}^{\pm.046}$   
    \\ \midrule

\multirow{21}{*}{\makecell[c]{KIT-ML}}  &  Real motion & 
 $0.424^{\pm.005}$ &
 $0.649^{\pm.006}$ &
 $0.779^{\pm.006}$ &
 $0.031^{\pm.004}$ &
 $2.788^{\pm.012}$ &
 $11.08^{\pm.097}$ &
  -
  \\ 
\cline{2-9} \vspace{-0.3cm} \\ 
~& TM2T \cite{guo2022tm2t}&
  $0.280^{\pm.005}$ &
  $0.463^{\pm.006}$ &
  $0.587^{\pm.005}$ &
  $3.599^{\pm.153}$ &
  $4.591^{\pm.026}$ &
  $9.473^{\pm.117}$ &
  $\textbf{3.292}^{\pm.081}$ \\
  
~& T2M \cite{guo2022generating}&
  $0.370^{\pm.005}$ &
  $0.569^{\pm.007}$ &
  $0.693^{\pm.007}$ &
  $2.770^{\pm.109}$ &
  $3.401^{\pm.008}$ &
  $10.91^{\pm.119}$ &
  $1.482^{\pm.065}$ \\

~& MDM \cite{tevet2022human}&
  $0.164^{\pm.004}$ &
  $0.291^{\pm.004}$ &
  $0.396^{\pm.004}$ &
  ${0.497}^{\pm.021}$ &
  $9.190^{\pm.022}$ &
  ${10.85}^{\pm.109}$ &
  ${1.907}^{\pm.214}$ \\

~& MLD \cite{chen2023executing} &
  ${0.390}^{\pm.003}$ &
  ${0.609}^{\pm.003}$ &
  ${0.734}^{\pm.002}$ &
  ${0.404}^{\pm.013}$ &
  ${3.204}^{\pm.010}$ &
  $10.80^{\pm.082}$ &
  ${2.192}^{\pm.079}$ \\

~& T2M-GPT \cite{zhang2023generating} &
    ${0.416}^{\pm.006}$ &
    ${0.627}^{\pm.006}$ &
    ${0.745}^{\pm.006}$ &
    ${0.514}^{\pm.029}$ &
    ${3.007}^{\pm.023}$ &
    ${10.92}^{\pm.108}$ &
    $1.570^{\pm.039}$ \\  

~& AttT2M \cite{zhong2023attt2m} &
    ${0.413}^{\pm.006}$ &
    ${0.632}^{\pm.006}$ &
    ${0.751}^{\pm.006}$ &
    ${0.870}^{\pm.039}$ &
    ${3.039}^{\pm.021}$ &
    ${10.96}^{\pm.123}$ &
    $\underline{2.281}^{\pm.047}$ \\ 

~& ParCo \cite{zou2024parco} &
    ${0.430}^{\pm.004}$ &
    ${0.649}^{\pm.007}$ &
    ${0.772}^{\pm.006}$ &
    ${0.453}^{\pm.027}$ &
    ${2.820}^{\pm.028}$ &
    ${10.95}^{\pm.094}$ &
    $1.245^{\pm.022}$ \\

~& ReMoDiffuse \cite{zhang2023remodiffuse} &
    ${0.427}^{\pm.014}$ &
    ${0.641}^{\pm.004}$ &
    ${0.765}^{\pm.055}$ &
    ${0.155}^{\pm.006}$ &
    ${2.814}^{\pm.012}$ &
    ${10.80}^{\pm.105}$ &
    $1.239^{\pm.028}$ \\ 

~& MMM \cite{pinyoanuntapong2024mmm} &
    ${0.404}^{\pm.005}$ &
    ${0.621}^{\pm.005}$ &
    ${0.744}^{\pm.004}$ &
    ${0.316}^{\pm.028}$ &
    ${2.977}^{\pm.019}$ &
    ${10.91}^{\pm.101}$ &
    $1.232^{\pm.039}$ \\ 

 ~& SALAD \cite{hong2025salad} &
    ${\underline{0.477}}^{\pm.006}$ &
    ${\textbf{0.711}}^{\pm.005}$ &
    ${\textbf{0.828}}^{\pm.005}$ &
    ${0.296}^{\pm.012}$ &
    ${\textbf{2.585}}^{\pm.016}$ &
    ${\textbf{11.097}}^{\pm.095}$ &
    $1.004^{\pm.040}$ \\

~& BAD \cite{hosseyni2025bad} &
    ${0.417}^{\pm.006}$ &
    ${0.631}^{\pm.006}$ &
    ${0.750}^{\pm.006}$ &
    ${0.221}^{\pm.012}$ &
    ${2.941}^{\pm.025}$ &
    ${11.00}^{\pm.100}$ &
    $1.170^{\pm.047}$ \\

 ~& BAMM \cite{pinyoanuntapong2024bamm} &
    ${0.438}^{\pm.009}$ &
    ${0.661}^{\pm.009}$ &
    ${0.788}^{\pm.005}$ &
    ${0.183}^{\pm.013}$ &
    ${2.723}^{\pm.026}$ &
    $\underline{11.01}^{\pm.094}$ &
    ${1.609}^{\pm.065}$ \\

 ~& MoMask \cite{guo2024momask} &
    ${0.433}^{\pm.007}$ &
    ${0.656}^{\pm.005}$ &
    ${0.781}^{\pm.005}$ &
    ${0.204}^{\pm.011}$ &
    ${2.779}^{\pm.022}$ &
    - &
    $1.131^{\pm.043}$ \\

 ~& Light-T2M \cite{zeng2025light} &
    ${0.444}^{\pm.006}$ &
    ${0.670}^{\pm.007}$ &
    ${0.794}^{\pm.005}$ &
    ${0.161}^{\pm.009}$ &
    ${2.746}^{\pm.016}$ &
    - &
    $1.005^{\pm.036}$ \\

 ~& MoGenTS \cite{yuan2024mogents} &
    ${0.445}^{\pm.006}$ &
    ${0.671}^{\pm.006}$ &
    ${0.797}^{\pm.005}$ &
    ${0.143}^{\pm.004}$ &
    ${2.711}^{\pm.024}$ &
    ${10.92}^{\pm.090}$ &
    - \\

 ~& LAMP \cite{li2024lamp} &
    ${\textbf{0.479}}^{\pm.006}$ &
    ${\underline{0.691}}^{\pm.005}$ &
    ${\underline{0.826}}^{\pm.005}$ &
    $\underline{0.141}^{\pm.013}$ &
    ${2.704}^{\pm.018}$ &
    ${10.929}^{\pm.101}$ &
    - \\

 ~& DisCoRD \cite{cho2025discord} &
    ${0.434}^{\pm.007}$ &
    ${0.657}^{\pm.005}$ &
    ${0.775}^{\pm.004}$ &
    ${0.169}^{\pm.010}$ &
    ${2.792}^{\pm.015}$ &
    - &
    $1.266^{\pm.046}$ \\

 ~& BiPO \cite{hong2026bipo} & 
    ${0.444}^{\pm.005}$ &
    ${0.674}^{\pm.006}$ &
    ${0.803}^{\pm.005}$ &
    ${0.164}^{\pm.008}$ &
    $\underline{2.658}^{\pm.015}$ &
    ${10.833}^{\pm.111}$ &
    $1.098^{\pm.047}$ \\

 ~& Motion Anything \cite{zhang2025motion} &
    ${0.449}^{\pm.007}$ &
    ${0.678}^{\pm.004}$ &
    ${0.802}^{\pm.006}$ &
    $\textbf{0.131}^{\pm.003}$ &
    ${2.705}^{\pm.024}$ &
    ${10.94}^{\pm.098}$ &
    ${1.374}^{\pm.069}$ \\   

\cline{2-9} \vspace{-0.3cm} \\ 
~& \textbf{ParTY (Ours)} &
    ${0.449}^{\pm.006}$ &
    ${0.680}^{\pm.007}$ &
    ${0.804}^{\pm.006}$ &
    ${0.155}^{\pm.014}$ &
    ${2.694}^{\pm.030}$ &
    ${11.21}^{\pm.082}$ &
    ${1.166}^{\pm.049}$ 
    \\ \bottomrule
\end{tabular}%
}
\label{tab:tables9}
\end{table*}

\end{document}